\documentclass[journal,transmag]{IEEEtran}
\usepackage{etoolbox}
\makeatletter
\pdfoutput=1
\patchcmd{\@makecaption}
{\scshape}
{}
{}
{}
{}
\patchcmd{\@makecaption}
  {\\}
  {.\ }
  {}
  {}
\makeatother

\usepackage{float}
\usepackage{multirow}
\usepackage[ruled,vlined]{algorithm2e}
\usepackage{cite}

\usepackage{mathrsfs}
\usepackage{xcolor}
\usepackage{float}
\usepackage{graphics}
\usepackage{booktabs}
\usepackage{threeparttable}
\usepackage{multirow}
\usepackage{subfigure}
\usepackage{makecell}
\usepackage{colortbl}  
\usepackage{xcolor}
\usepackage{array}

\ifCLASSINFOpdf
\usepackage[pdftex]{graphicx}
\else  
\fi

\usepackage[cmex10]{amsmath}
\usepackage{enumerate}
\usepackage{algorithmic}

\usepackage{url}

\begin{document}
\title{FasterX: Real-Time Object Detection Based on Edge GPUs for UAV Applications}
\author{Wei~Zhou,
        Xuanlin~Min,
        Rui~Hu,
        Yiwen~Long,
        Huan~Luo,
	   and JunYi
\thanks{}
\thanks{W. Zhou, X. Min, R. Hu, Y. Long, H. Luo, and J. Yi are with the School of Intelligent Technology and Engineering, Chongqing University of Science and Technology, Chongqing 401331, China (e-mail: zhouw@cqust.edu.cn).}
}

\maketitle
\begin{abstract}
Real-time object detection on Unmanned Aerial Vehicles (UAVs) is a challenging issue due to the limited computing resources of edge GPU devices as Internet of Things (IoT) nodes. To solve this problem, in this paper, we propose a novel lightweight deep learning architectures named FasterX based on YOLOX model for real-time object detection on edge GPU. First, we design an effective and lightweight PixSF head to replace the original head of YOLOX to better detect small objects, which can be further embedded in the depthwise separable convolution (DS Conv) to achieve a lighter head. Then, a slimmer structure in the Neck layer termed as SlimFPN is developed to reduce parameters of the network, which is a trade-off between accuracy and speed. Furthermore, we embed attention module in the Head layer to improve the feature extraction effect of the prediction head. Meanwhile, we also improve the label assignment strategy and loss function to alleviate category imbalance and box optimization problems of the UAV dataset. Finally, auxiliary heads are presented for online distillation to improve the ability of position embedding and feature extraction in PixSF head. The performance of our lightweight models are validated experimentally on the NVIDIA Jetson NX and Jetson Nano GPU embedded platforms.
Extensive experiments show that FasterX models achieve better trade-off between accuracy and latency on VisDrone2021 dataset compared to state-of-the-art models.
\end{abstract}
\begin{IEEEkeywords}
Lightweight, prediction head, distillation, edge GPU, UAV.
\end{IEEEkeywords}

\section{Introduction}
\IEEEPARstart{U}{nmanned} aerial vehicles (UAVs) have been considered as an efficient Internet of Things (IoT) node for large-scale environment sensing and monitoring \cite{IOT1,IOT2}, which is widely used in urban \cite{1}, agricultural \cite{2}, surveillance \cite{3,4}, and other tasks. Visual object detection is a hot topic in UAV application that can locate and classify all objects in UAV photography, such as pedestrians, cars, bicycles, and so on.
In recent years, object detection based on deep learning has made significant progress both in accuracy as well as efficiency, and many excellent networks have been proposed. The two-stage model \cite{5,6,7} usually achieves high accuracy but performs poorly in efficiency, which is difficult to perform on UAV platform with limited computing resources. Recently, one-stage model based on YOLO is widely used in embedded system \cite{8,9,10}. However, the anchor-based YOLO models, it does not solve the following problems: 1) the anchor needs to be carefully and manually redesigned to adapt to the anchor distribution of different data sets; 2) the imbalance between positive samples and negative samples. Meanwhile, it is still a challenging task to balance detection accuracy and real-time requirements.

\begin{figure}
\centering
\includegraphics[width=0.5\textwidth]{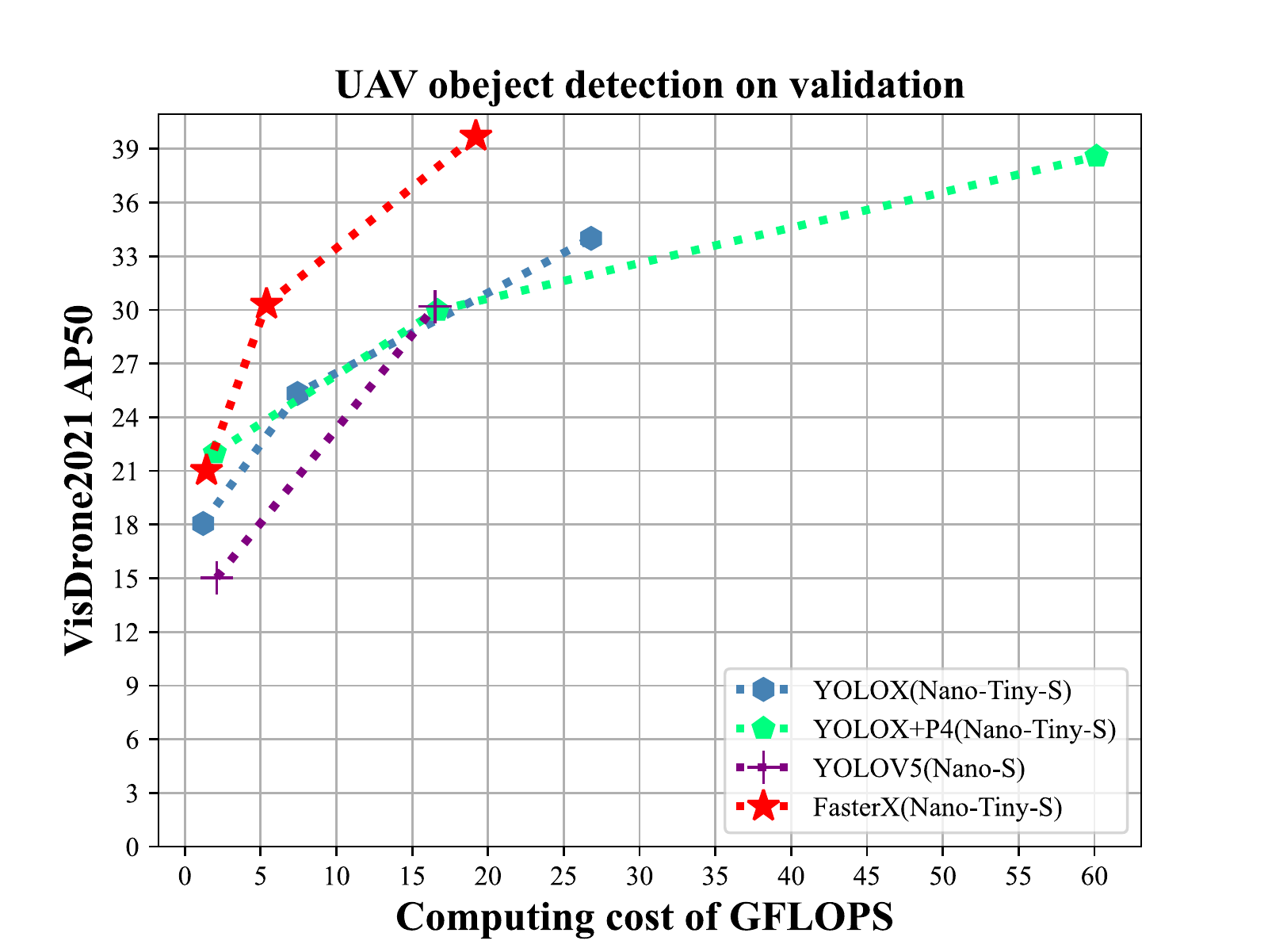}
\caption{Giga Floating-point Operations Per Second (GFLOPs) and accuracy (AP50) on VisDrone2021 benchmark dataset. Our FasterX is much better than YOLOV5, YOLOX, and YOLOX+P4. Noting that Nano-Tiny-S is in order from the bottom to the top of the curve in YOLOX, and YOLOV5 inluding Nano-S. Such performance is very competitive in UAV applications. Details are given in Section IV.}
	\label{Fig.system}
\end{figure}

\begin{figure}
\centering
\includegraphics[width=0.5\textwidth]{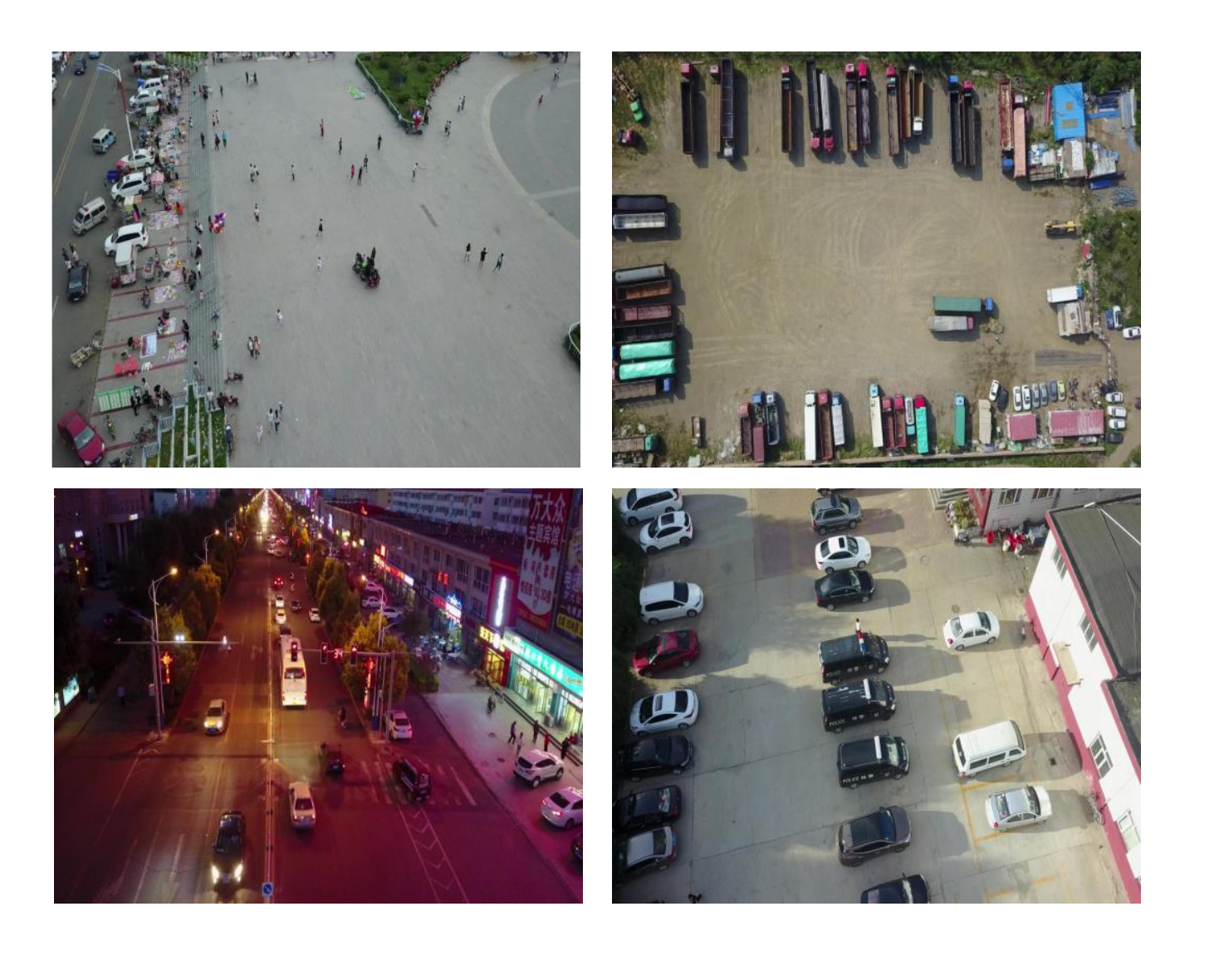}
\caption{Some samples in Visdrone2021 dataset \cite{17}, which illustrate the large and dense number of small objects and complex background in UAV dataset.}
\label{Fig.datasets}
\end{figure}

To overcome this problem, many researchers have been committed to develop efficient detector architectures, such as FCOS \cite{11} and FSAF \cite{12}. The lightweight model based on anchor-free, Nanodet\cite{13}, YOLOX-Nano, YOLOX-Tiny, and YOLOX-S \cite{14} are also typical representatives. These detectors conduct a model evaluation based on COCO, VOC, etc, which promotes the development of the object detection field.
However, there are two practical issues for the application of UAV detection on the edge devices: 1) compared with COCO dataset \cite{15} and VOC dataset \cite{16}, UAV dataset have specific problems such as large changes in object scale and a large number of small objects, which is intuitively illustrated by some cases in Fig.~\ref{Fig.datasets}; 2) there is a certain hardware bottleneck in UAV real-time edge monitoring. Hence, this is an irreconcilable contradiction between detection accuracy of small objects and model inference speed.
For UAV, current the simple and efficient methods to improve small object accuracy can be summarized as two points: 1) increasing the input image resolution to enlarge the object; 2) without changing the resolution of the input image, an additional small object detection head is supplemented by adding a large-resolution feature map, which is an amplification strategy on the feature map \cite{18}.
No matter how to improve the accuracy, the above two tricks will lead to a sharp rise in the GFLOPS cost of the model, seriously affecting the inference and not conducive to the real-time edge detection. Hence, these methods cannot be simply applied to lightweight model.

Recently, relevant work has been put forward in succession for the embedded deployment of UAV \cite{19,20}. Among them, the model compression method proposed by Zhang et al. \cite{20} has been widely validated on Visdrone2021 dataset \cite{17} to show its effectiveness, which can be used as one of the optimization schemes.
Unlike model compression, this paper focuses on the optimization of small objects and detector architecture, to realize the real-time inference performance on edge GPU devices for UAV.
Inspired by PP-PicoDet \cite{21}, NanoDet-Plus \cite{13}, ESPCN \cite{22} and YOLOX \cite{14}, we present an edge GPU friendly anchor-free detector FasterX for UAV detection, which is much more accurate and lighter than lightweight detector YOLOX. With adding more prediction heads, our FasterX still maintain the advantages of accuracy and efficiency on parameters, GFLOPS cost and inference time.
In summary, the main contributions of this paper are as follows:
\begin{itemize}
\item In this paper, a novel lightweight PixSF head is proposed to greatly alleviate the reasoning puzzle caused by the Head layer, where an additional detection head P4 is added. Furthermore, the PixSF head is constructed by a novel position encoder-decoder (Fcous \& Pixel Shuffling). The PixSF head is not only effective on UAV dataset Visdrone2021, but also experimentally proved on the VOC2012 benchmark dataset.

\item We develop a slimmer structure to replace PAFPN structure in the Neck layer, named as SlimFPN. The SlimFPN removes the upper sampling structure of PAFPN and unifies the input channel numbers of all branches of the Neck by Ghost module\cite{24}, which significantly reduces parameters of the network.

\item We employ dynamic label assignment strategy SimOTA \cite{42,14} to optimize training process to obtain global optimization result. Meanwhile, we adopt the weighted sum of Focal Loss (FL) \cite{25} and complete intersection over union (CIoU) loss \cite{26} to alleviate category imbalance and aspect ratio imbalance of the ground truth (GT) box.

\item We present a novel auxiliary heads for online distillation to improve the ability of position encoding and feature extraction in PixSF head.
\end{itemize}

The rest of this article is organized as follows: Section II covers the related work while Section III discusses the proposed FasterX. In Section IV, experimental results and discussions are provided. Section V summarizes the full text and plans for future work.

\begin{figure*}
\centering
\includegraphics[width=0.9\textwidth]{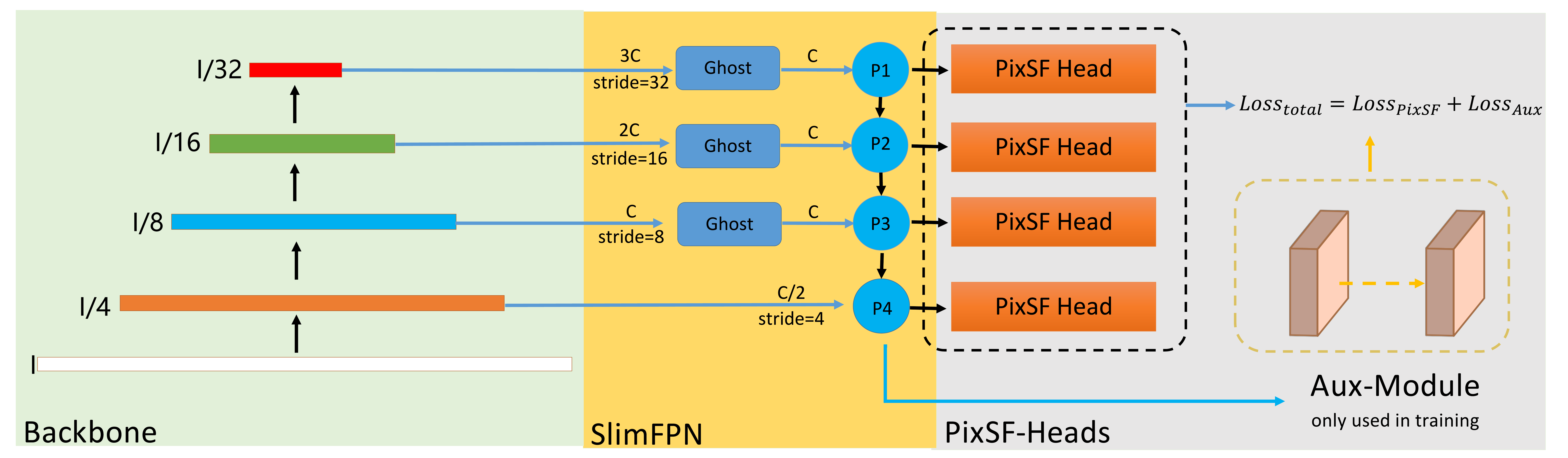}
\caption{Framework of FasterX. The Backbone is CSPDarknet53, which outputs P1-P4 feature maps to the Neck layer with strides (32,16,8,4). In SlimFPN, parameter reduction for Neck inputs(P1, P2 and P3) with large number of channel is achieved through Ghost module \cite{24}. In Head layer, four lightweight PixSF heads are proposed to balance accuracy and speed. In addition, Aux-Module is designed to obtain a better label assignment for PixSF Head.}
\label{Fig.framework}
\end{figure*}

\section{Related Work}
In the development of deep learning, lightweight has always been one of the tasks of neural network optimization. In the field of object detection, great progress has been made in lightweight components. This section will elaborate on the object detection and its lightweight technique.

The detection networks using CNN as the feature extractor can be divided into many types. For example, anchor-based and anchor-free can be categorized according to different label assignment strategies. According to different stages classification, object detection can be divided into one-stage and two-stage detectors.

From the perspective of components, the current mainstream technology generally consists of three parts. One part is based on the CNN backbone, which is used for image feature extraction, the other part is to design Neck layer for feature fusion, and finally, based on the feature assignment of Neck layer, the detection head is used for category and box prediction. Next, we will introduce the three components of object detection and the related lightweight technologies.

\textbf{Backbone:} The backbone of object detection mainly composes of CNN network, such as VGG, ResNet, DenseNet, Swin Transformer etc \cite{27,28,29,30}. Lightweight design of the network includes Mobilenet, GhostNet, and ShuffleNet \cite{24,31,32}. Although these works have accelerated the model by reducing parameters and GFLOPS, recent studies have shown that evaluating only these theoretical indicators may lead to suboptimal results. Some researchers propose that it is a feasible scheme to directly use edge devices for practical validation. Repvgg \cite{41} uses structural re-parameterization to improve the model inference speed, and proves that only using Flops to measure the real speed of different architectures is inappropriate. YOLO-ReT\cite{9} improves convolution module by building a Neck structure more conducive to edge device inference, which measures real speed by inferring directly on the target device.

\textbf{Neck:} The Neck layer is designed to better extract features from the backbone. Neck layer can reprocess the feature maps extracted by the backbone network at different stages, and reasonably uses different receptive fields for sample point assignment. Generally the Neck layer consists of several bottom-up and top-down paths.

Improved path aggregation blocks based on FPN include FPN\cite{7}, PANet\cite{33}, NAS-FPN\cite{34}, BiFPN\cite{35}, ASFF\cite{36}, and SFAM\cite{37}. Path aggregation block is very effective, but its multi-branch structure and the repetition of various upsampling and downsampling and splicing operations will reduce the inference speed. Recently, there have been many studies on the lightweight Neck. YOLOF \cite{38} proposed dilated encoder and uniform matching to replace FPN. NanoDet\cite{13} completely removes all convolutions in PANet, and uses interpolation to complete up sampling and down sampling. PPYOLOE \cite{39} and YOLOV7 \cite{40} all introduced re-parameterized convolution \cite{41} and applied it to the Neck for improvement of inference.

\textbf{Head:} The head layer is designed for extracting feature maps from the Neck layer, handling position regression and classification tasks.
In recent years, many improved methods for head layer mainly focus on regression and classification task \cite{43,44,45}.
A common optimization method is to divide the head layer into different streams according to regression and classification, and performed separately feature extraction \cite{14}. This method can alleviate the contradiction between regression features and classification features. However, this method will bring inference latency on edge devices.
Recently, Transformer methods have been a hot topic in computer vision \cite{18,46}.
Several studies have explored the effects of using transformer to improve the prediction accuracy in the head layer.
Unlike convolution operator, transformer is a non-intensive operation on edge GPU, which resulting in underutilization of computational resources. Meanwhile, a large number of element-wise operations may significantly slow down the inference speed.
The requirements of computational and storage make it difficult to deploy on edge GPU devices.

\section{Proposed Models}

\subsection{YOLOX}
\label{sec:3}
YOLOX is designed based on the anchor-free framework, abandoning the anchor-based strategy of YOLO series (YOLOv2-v5). Because it has achieved the most advanced results in the current YOLO series on COCO datasets, this paper chooses it as the baseline. A total of six different models with different network width and depth setting are available for YOLOX, including YOLOX-Nano, YOLOX-Tiny, YOLOX-S, YOLOX-M, YOLOX-L and YOLOX-X. Among them, YOLOX-Nano, YOLOX-Tiny, and YOLOX-S belong to the lightweight family of YOLOX and are suitable for deployment in embedded devices.

\subsection{FasterX}
\label{FasterX}
The framework of FasterX is shown in Fig.~\ref{Fig.framework}. The main part of the FasterX is consists of PixSF head, SlimFPN, and auxiliary head. The proposed FasterX achieves efficient and real-time deep learning-based object detection on Jetson NX and Jetson Nano GPU embedded platforms. Next, we will describe this design in detail.

\begin{figure}
\raggedright
\includegraphics[width=0.5\textwidth]{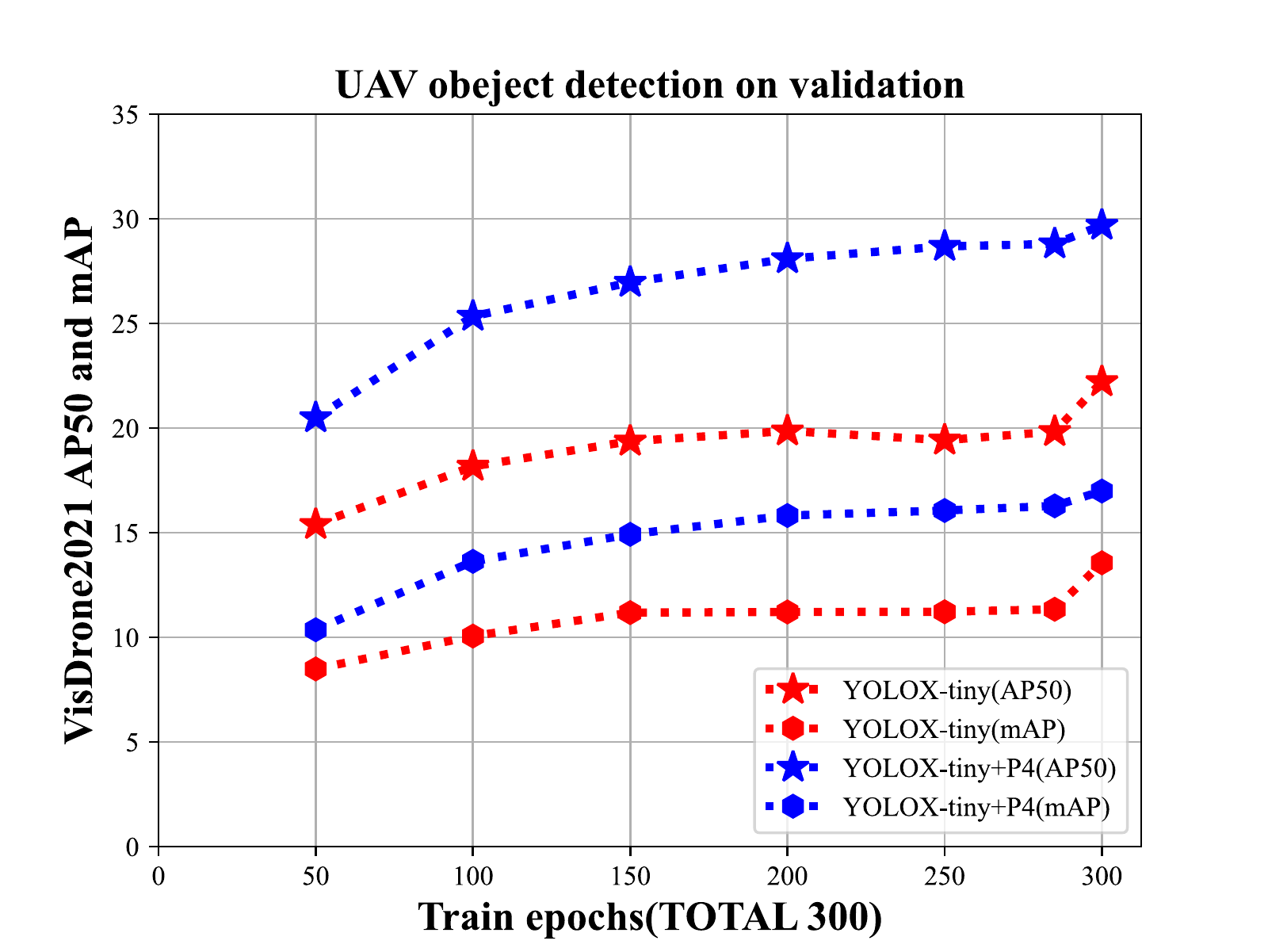}
\caption{Training curves for detectors with 4-head (add P4 head) and 3-head structure. We evaluate the mAP and AP50 on Visdrone2021 every 50 epochs. It is obvious that the 4-head converges much faster than the 3-head and also achieves better result.}
\label{4head}
\end{figure}

\textbf{Lightweight PixSF head:} For UAV object detection application, the objects are relatively small in the overall image scene (small object $<32\times32$ pixels, medium object from $32\times32$ pixels to $96\times96$ pixels, and large object $>96\times96$ pixels, defined by COCO dataset). For example, in VisDrone2021 dataset, more than 60\% objects are small object such as pedestrian, bicycle, motor and car.
Although many deep learning-based object detection methods can achieve high accuracy, these methods perform poorly in UAV dataset due to the small object and complex backgrounds.
Recently, many studies and reviews have been undertaken to explore using an additional detection head (P4 head) to improve the detection accuracy on small objects \cite{18,11}. The 4-head structure for the prediction head can alleviate the impact of the change of object scale and reduce the difficulty of optimization for small object detection network.
As shown in Fig.~\ref{4head}, we compare three benchmark models of the lightweight series YOLOX under the 4-head structure. The mean average precision (mAP) and average precision at AP50 of 4-head YOLOX-Tiny are far better than the 3-head structure.

Although the 4-head structure is very effective, the parameters and GFLOPS increase significantly on the GPU embedded device. Especially, the GFLOPS increases seriously because of the further expansion of the multi-branch structure, which is not conducive to GPU parallel computing. Table~\ref{latency} shows the latency results on the Jetson NX and Nano platforms. Taking YOLOX-S as an example, the inference delay of YOLOX with a 4-head structure increases 85\% and 353\% on the Jetson NX and Nano platforms, respectively, compared with the native YOLOX.
To address this problem, in this paper, a lightweight PixSF head is designed based on a 4-head structure, which has a trade-off between accuracy and speed.

\begin{table}[!t]
\caption{The latency comparisons between 4-head and 3-head structures on Jetson NX and Jetson Nano platforms.}
 \label{table.dataset}
 \centering   
 \begin{tabular}{|l|c|c|c|}
\hline
   Model & \thead{Params\\(M)} &GFLOPs & \thead{Latency\\Nano\quad NX} \\
\hline\hline
  YOLOX-S&9.0&26.8&52.52ms\quad14.70ms\\
\multirow{1}{*}{YOLOX-S+P4}
  &9.69&60.14&\textbf{238.22ms}\quad\textbf{27.26ms}\\
\hline\hline
  YOLOX-Tiny&5.06&7.43&38.89ms\quad10.68ms\\
\multirow{1}{*}{YOLOX-Tiny+P4}
 &5.45&16.67&\textbf{79.85ms}\quad\textbf{12.89ms}\\
\hline\hline
YOLOX-Nano&0.91&1.21&27.57ms\quad8.91ms\\
\multirow{1}{*}{YOLOX-Nano+P4}
&0.94&1.98&\textbf{55.25ms}\quad\textbf{10.96ms}\\
\hline
\end{tabular}
 \label{latency}
\end{table}

\begin{figure*}
\centering
\includegraphics[width=0.9\textwidth]{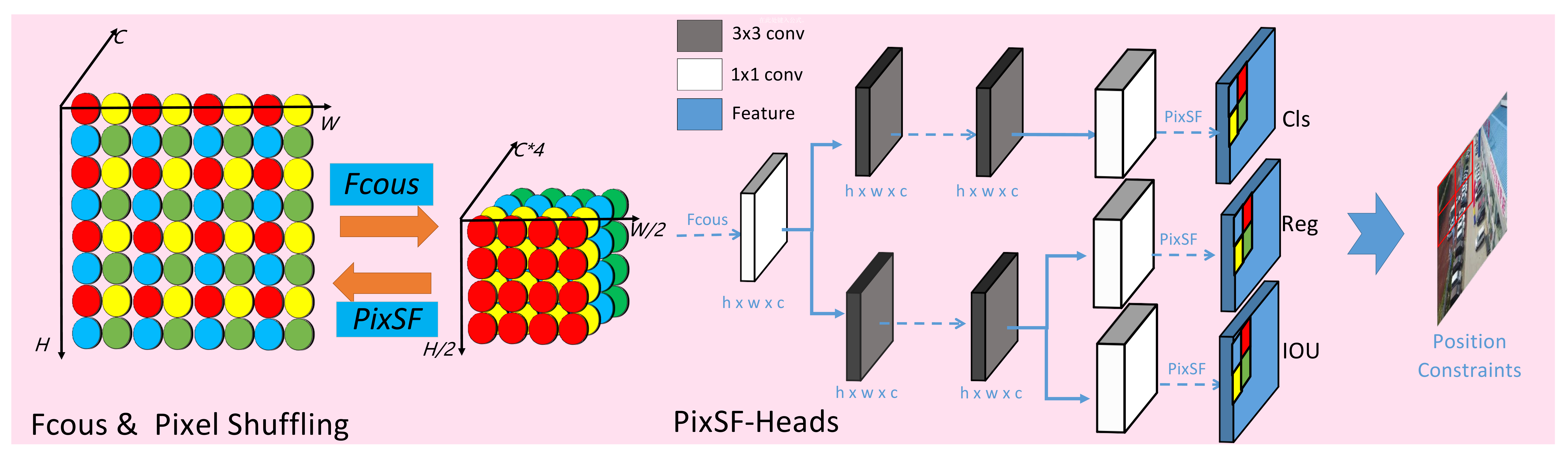}
\caption{Framework of PixSF head. Focus and Pixel Shuffling are one-to-one encoding and decoding operations, where Focus is responsible for embedding the position in channel-wise, and Pixel Shuffling is responsible for decoding and restoring the position information spatially. Pixsf heads uses this pair of key operations to construct a new head-level paradigm in combination with YOLO location constraints.}
\label{Fig.pixsfhead}
\end{figure*}

In the field of super-resolution reconstruction, the loss function is usually constructed for pixel-level supervision to realize the mapping from Low-resolution image (LR) to High-resolution image (HR), in which the pixel-level loss function (L2 loss) and convolution operation implicitly contain the location information \cite{47,48,22}. Similarly, position constraints also exist in object detection. When predicting the confidence of position regression, the position prediction needs to be decoded to the input coordinate range to adapt loss supervision. In YOLO detectors, the position decode can be formulated as follows:
\begin{align}\label{eq20}
[t_{x},t_{y}]&=(Convs(\textbf{F})+ Grid\_coord)*Stride  \nonumber\\ [t_{w},t_{h}]&=e^{Convs(\textbf{F})}*Stride
\end{align}
where $t_{x}$ and $t_{y}$ denote the offset of the center point from the upper left corner of the grid cell, and $t_{w}$ and $t_{h}$ represent the width and height factors of the object, respectively. These values are within the scale range of original map and need to be linked by $Convs(\textbf{F})$ and the stride of the current feature map relative to original map. $Grid\_coord$ represents the upper-left coordinate of the grid cell. Further, $\textbf{F}$ represents the input of the head layer, and $Convs(\cdot)$ represents the convolution operation in the head layer.

In summary, the formula (\ref{eq20}) can be considered as a feature position embedding by mapping the encoded feature to the fixed position of original map and combing with  grid and downsampling multiples. Inspired by this, in this paper, a position encode-decoder is designed by using the position embedding in the head layer, named PixSF head.

As shown in Fig.~\ref{Fig.pixsfhead}, the encoder uses Focus to concatenate pixel-patch along channel-wise in the HR feature. In this way, the channel of each pixel point contains patch position information (upper left corner, upper right corner, lower left corner, lower right corner). Furthermore, in order to reduce the encoding cost and build the position relationship between the pixel-patchs, we use a simple $1\times1$ convolution for feature extraction and dimension reduction.
Then, we input it into the decoupled head for regression and classification tasks, respectively.
In the decoder part, we restore the extracted features of the encoder to $C \times rH \times rW$ dimensions by using pixelshuffle operation. Specifically, the $1\times1$ convolution with a depth of $r^{2}*C$ is used to map the graph of $C \times H \times W$  to $r^{2}*C \times H \times W$, where $r$ is set to be 2. The encode-decoder can be described in the following way:
\begin{align}\label{eq101}
EC(X)&=\phi(\omega_{1}(FCOUS(X_{[C,H,W]})))  \nonumber\\ DC(X)&=Pixelshuffle(EC(X_{[C,H,W]})) \nonumber\\
FCOUS(X)&=X_{[C,H,W]} \rightarrow X_{[r^2*C,H/r,W/r]} \nonumber\\
Pixelshuffle(X)&=\phi(\omega(FCOUS(X)))_{[C,H,W]}
\end{align}
where $EC(\cdot)$ represents the encoder, and $DC(\cdot)$ represents the decoder. $X_{[C,H,W]}$ is input $X$ with size of $C \times H \times W$ and $r$ is the multiple factor, noting that $r$ needs to be divisible by $H$ or $W$. The activation function $\phi$ is applied element-wise and $\omega$ is the parameter included in the whole decoupling layer.

\textbf{SlimFPN:}
Anchor-free object detector mainly adopt feature pyramid network (FPN) for multi-scale prediction, which makes it possible to extract multi-level object feature to improve online feature selection capability. In the path aggregation network (PANet) of YOLOX, a bottom-up structure PAFPN based on FPN is designed to complement localization information, which enables FPN has a bottom-up gradient update. In the PAFPN structure, the input and output of the Neck layer are the same as the output of the backbone at different stages. The PAFPN structure can be formulated as
\begin{equation}
\begin{aligned}
&P4^{'}=DS_{2X}(P4) \\
&P3^{'}=f(DS_{2X}(Cat(P4^{'},P3))) \\
&P2^{'}=f(DS_{2X}(Cat(P3^{'},P2))) \\
&P1^{'}=f(DS_{2X}(Cat(P2^{'},P1)))
\end{aligned}
\end{equation}
where P4 is the lowest layer of the FPN, $DS_{2X}$ represents 2 times downsampling, $f$ is the fusion operator, $Cat$ is the channel cascade.

PANet structure can propagate localization features to high semantic feature layer by combining the downsampling the upper level feature map and the present level feature map. However, from the point of view of gradient propagation, the P4 prediction layer, which is benefit for small object detection, has been calculated the loss before the PANet structure. Hence, P4 head cannot share the advantages of PANet structure on the gradient, but P1$'$ head, which is benefit for big objects, can fully share the advantages of PANet structure.

Consider that the PANet with a large number of channels is structurally multi-branches and result in delayed inference on edge GPU devices, in this paper, we design a  lightweight SlimFPN structure without PANet as the Neck layer.
SlimFPN forms the multilevel feature map by intercepting the output of each stage in backbone. The extraction parameters of network increases with the depth of the network, which leads the increase of the feature channels. A large number of channels will increase the computational cost of edge GPU devices. 
Hence, in SlimFPN, channel scaling is performed on large parameters branches of features to reduce inference speed. As shown in the SlimFPN part of Fig.~\ref{Fig.framework}, since the backbone channels increase with depth at each stage, the input $C$ of P3 is used as the basis to represent other levels of input. Considering that balancing channels will inevitably result in loss of the information, parameter reduction is achieved through Ghost module \cite{24} to increase the spatial extraction ability to alleviate the loss of the information.

Compared to PANet, the proposed SlimFPN may result in some loss of accuracy, but the influence is insignificant. This is because most of the UAV objects are considered to be small objects. On the one hand, it benefits from the powerful performance of FPN in feature fusion. The top-down structure can ensure that the deep semantic information is transferred to the shallow feature maps, which can provide semantic support for small objects. On the other hand, the bottom-up structure of PANet is mainly aimed at large objects and does not play a supervisory role in P4 head. Therefore, SlimFPN will lead to the decline of large object accuracy but it can improve the inference speed in three models, which is a trade-off on accuracy and speed.

\textbf{Head with attention:} In the traditional head layer in object detection, regression and classification are always in the final stage, where most of the weights are shared between the positioning and regression tasks. Double head RCNN \cite{43} firstly proposes a double head layer, where different convolution are adopted to extract streams for regression and classification tasks to minimize the feature parts shared by regression and classification. Since then, this idea has been used in the recently popular one-stage detectors \cite{14,21,38}. In order to further improve the extraction ability in head, PPYOLOE \cite{39} and TOOD \cite{50} try to use the attention mechanism to enhance feature extraction ability, and the effectiveness of the attention module integrated in head is proved in COCO dataset.

\begin{figure}
\centering
\includegraphics[width=0.45\textwidth]{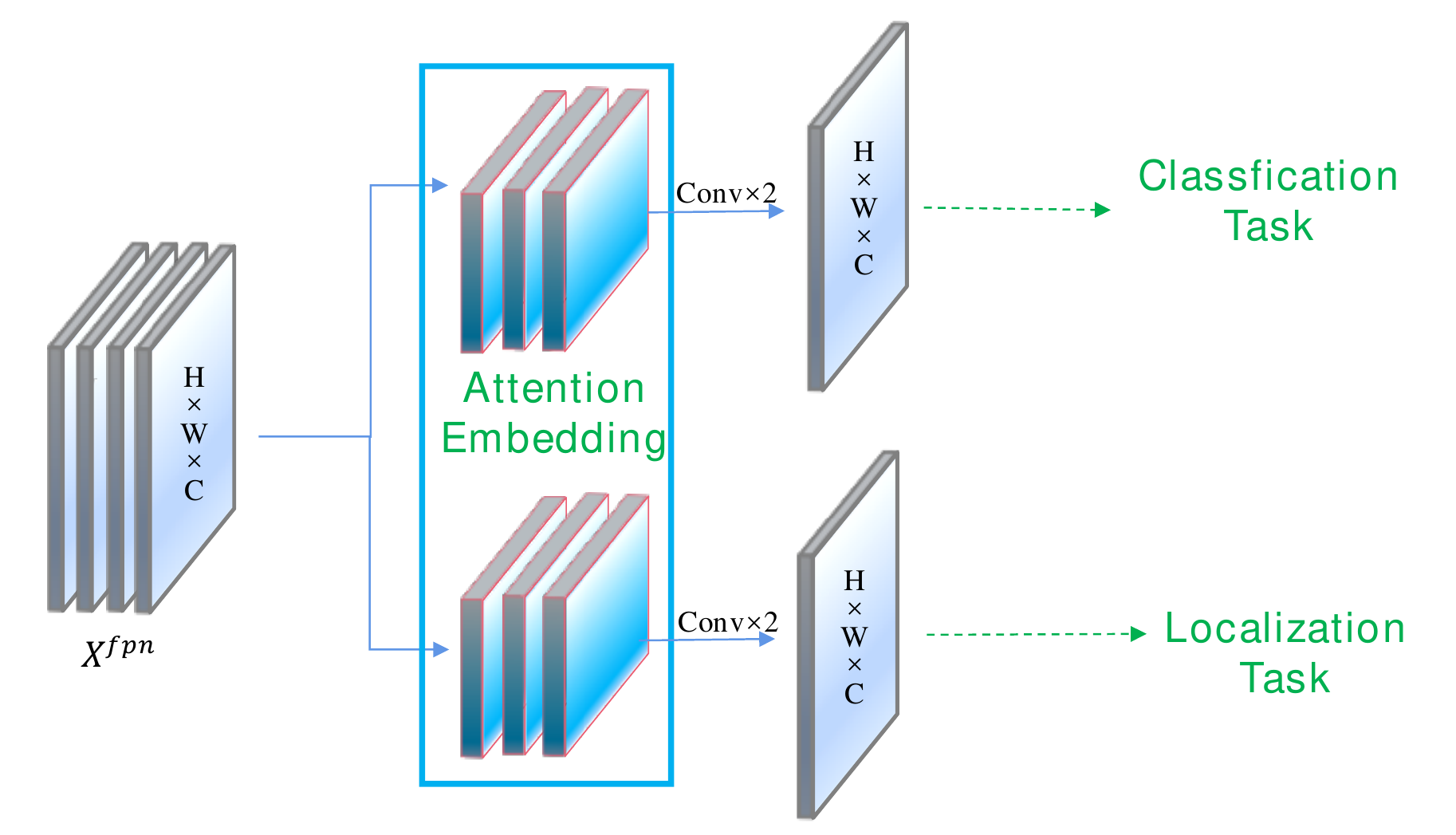}
\caption{Network structure of the attention mechanism in FasterX head.}
\label{Fig.attention}
\end{figure}

In the images of UAV, the large coverage area always contains complex and diverse backgrounds. To enhance feature extraction ability of the head layer, we adopt attention mechanism to focus on the interest area. In this paper, convolutional block attention module (CBAM) is employed to improve the feature representation of Head layer due to the abundant attention feature maps on channel and spatial.
Meanwhile, CBAM is a lightweight module that can be easily embedded into head layer to improve detection performance, as shown in Fig.~\ref{Fig.attention}.

\begin{figure*}
\centering
\includegraphics[width=1\textwidth]{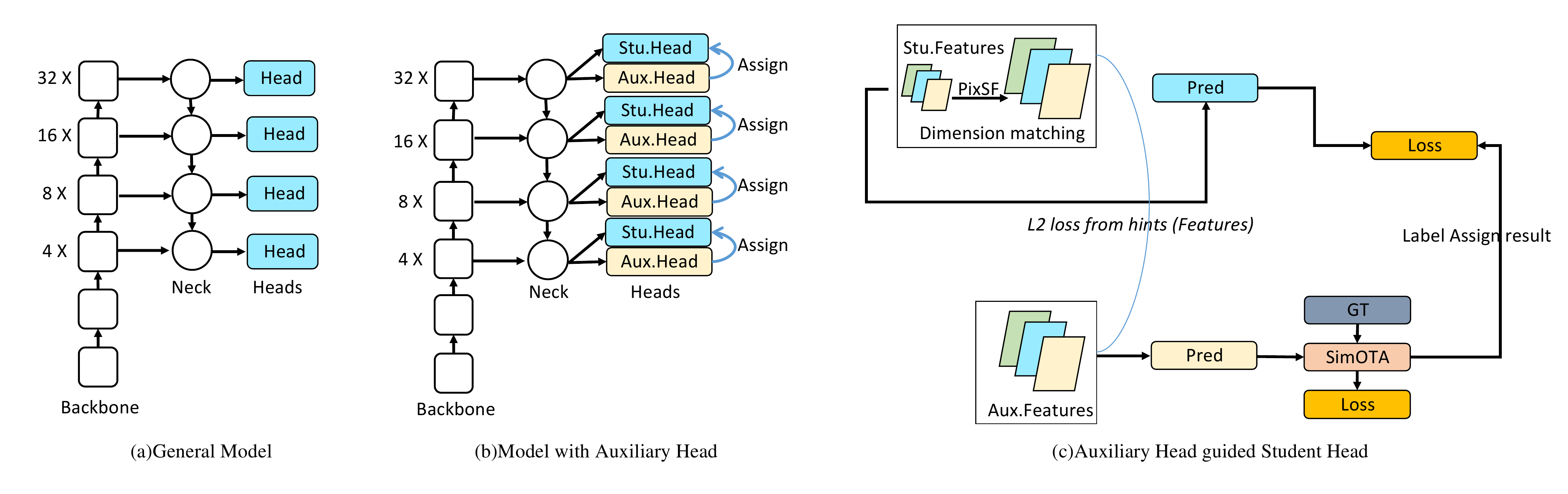}
\caption{Architecture diagram of the online distillation with the auxiliary head. (a) is the common object detection framework including Backbone, Neck and Heads. (b) is the auxiliary model, including Stu.Head and Aux.Head, where the Stu.Head is the PixSF head.
The auxiliary distillation process is shown in (c), Aux.Head is used to guide SimOTA to assign label assignment results for Stu.head. In addition, the features distance between two heads is optimized by L2 loss.}
\label{Fig.auxiliary}
\end{figure*}

\textbf{Label Assignment Strategy and Loss:} The label assignment of positive and negative samples has a crucial impact for the detector. Label assignment is assigned
Retinanet \cite{25}, Faster RCNN \cite{7} directly divides positive samples and negative samples through the intersection over union (IOU) threshold between the anchor template and ground truth (GT). YOLOV5 \cite{23} uses the aspect ratio between the Anchor template and GT to sample positively. FCOS \cite{11} takes anchors in the central area of GT as positive samples, and ATSS \cite{54} determines samples according to the statistical characteristics of the nearest anchors around GT. The above label assignment strategy is unchanged in the process of global training. In order to obtain the globally optimal assignment results, we use SimOTA dynamic label assignment strategy to optimize the training process.

SimOTA first determines the candidate region through a central prior, then calculates the IOU value between the GT and the candidate box in the candidate region, and finally obtains the dynamic parameter $K$ by summing the IOU. Furthermore, SimOTA uses the cost matrix to assign $K$ candidate boxes to each GT for network training, where the cost matrix is obtained by directly calculating losses in all candidate boxes and GT. The original SimOTA uses the weighted sum of CE loss and IOU loss to calculate the cost matrix, which is which can be expressed as:
\begin{equation}
CostMatrix=ClsLoss_{BCE}+ \alpha * RegLoss_{IOU}
\end{equation}

In the UAV scene, the categories of sample and the aspect ratio of the GT box are often unbalanced, respectively. Taking VisDrone2021 dataset as an example, the sample of pedestrian is 79 337, but the sample of tricycle is 4 812. Meanwhile, the aspect ratio of the GT box is also unbalanced. The range of aspect ratio (Max side / Min side ) is broad, from 1 to an upper limit of near 5. However, most of the GT box is between 1 and 2.5. In such cases, the original label assignment loss of YOLOX cannot handle handle it well.
To solve this problem, for the category imbalance, we use Focal loss to make the network pay more attention to the less-sample categories. Meanwhile, for the aspect ratio imbalance, we use CIOU loss replace IOU loss, which can introduce aspect ratio supervision information to fit the GT BOX. Hence, the Focal loss can be expressed as
\begin{equation}
CostMatrix=ClsLoss_{Focal}+ \alpha * RegLoss_{CIoU}
\end{equation}
\begin{equation}
ClsLoss_{Focal}=|y-\overline{y}|^{\gamma}*(\alpha_{1}*y+(1-\alpha_{1})*(1-y))
\end{equation}
where $\alpha$ is balance factor for CostMatrix and $|y-\overline{y}|$ means distance between predicted value and label. $\gamma$ and $\alpha$ are hyperparameters for adjusting unbalanced class. For regression part, $RegLoss_{CIoU}$ can be defined as:
\begin{equation}\label{eq201}
RegLoss_{CIoU}=1-IOU+\frac{distance(b,b^{gt})^{2}}{c^{2}}+\alpha_{2} v
\end{equation}
\begin{equation}\label{eq201}
v=\frac{4}{\pi^{2}}+(arctan\frac{w^{gt}}{h^{gt}}-arctan\frac{w}{h})^{2}
\end{equation}
where $\alpha_{2}$ also is balance factor. Then, $v$ is used to measure the similarity of aspect ratio and distance($\cdot$) compute distance between the central points of prediction box and GT box.

\textbf{Auxiliary Head for Online distillation:} In this paper, we use position constraints to design the encode-decoder PixSF head to improve the detection of small objects. Meanwhile, we use improved SimOTA method to train the network.
However, for lightweight network, dynamic matching and position encoding and decoding method may not obtain good results because the model is initialized randomly at the beginning \cite{55}.

To address this issue, in the lightweight PixSF head, we design an auxiliary head for online distillation. As shown in Fig.~\ref{Fig.auxiliary}, we design two streams (Stu.Head and Aux.Head) at the head level, where Stu.Head (that is the PixSF head) is the head layer of the original structure and is the distilled layer.
Aux.Head is designed to guide SimOTA to assign label assignment results for Stu.head (PixSF heads). As shown in Fig.~\ref{Fig.auxiliary} (c), the label assignment results of SimOTA guided by Aux.head is used to replace the original label assignment results of the Stu.Head, to improve the overall expressive ability of Stu.Head.
In addition, we add an additional pixel-level alignment task to enhance the ability of Stu.Head to learn from Aux.Head.
This process can be summarized as the following:
\begin{equation}\label{eq201}
\begin{split}
Loss_{total}=Loss_{PixSF}+Loss_{Aux}+\\
\lambda \cdot \parallel F_{PixSF}-F_{Aux} \parallel^{2}_{2}
\end{split}
\end{equation}
where $\parallel F_{PixSF}-F_{Aux} \parallel^{2}_{2}$ is used to compute distance between features on PixSF head and Aux.head.

It is worth noting that the auxiliary head is only performed during the training process without inference cost.

\label{sec:9}
\begin{table*}[!t]  \caption{Comparison with the state-of-the-art one-stage detectors on Visdrone2021 dataset.}
 \centering
 \begin{tabular}{c|c|c|c|c|c|c|c}
\toprule[1pt]
   Model & Size & Params(M) &GFLOPs &mAP(val)&AP50(val) &mAP(test) &AP50(test) \\
\midrule[1pt]
  YOLOv3-Tiny    & $832 \times838$ & 8.7  & 21.82 & $11.00\%$ & $23.40\%$ & - & - \\
  SlimYOLOv3-SPP3-50 & $832 \times838$ & 20.8 & 122 & $25.80\%$ & $45.90\%$ & - & - \\
  SlimYOLOv3-SPP3-90 & $832 \times838$ & 8.0 & 39.89 & $23.90\%$& $36.90\%$ & - & - \\
  SlimYOLOv3-SPP3-95 & $832 \times838$ & 5.1 & 26.29 & $21.20\%$    & $36.10\%$ & - & - \\
  \midrule
  YOLOv5-S    & $640\times640$ & 7.2  & 16.5 & $15.03\%$ & $30.19\%$ & $13.22\%$ & $25.86\%$ \\
  YOLOv5-Nano & $416\times416$ & 1.9 & 1.8 & $6.61\%$ & $14.60\%$ & $5.65\%$ & $13.03\%$ \\
  YOLOv5-Nano & $448\times448$ & 1.9 & 2.1 & $6.75\%$ & $15.03\%$ & $5.90\%$ & $13.67\%$ \\
\midrule
  YOLOX-S    & $640\times640$ & 9.0  & 26.8 & $19.80\%$  & $34.00\%$ & $16.86\%$ & $29.57\%$ \\
  YOLOX-Tiny &  $448\times448$ & 5.06 & 7.43 & $14.37\%$ & $25.33\%$ & $13.23\%$ & $22.67\%$ \\
  YOLOX-Nano &  $448\times448$ & 0.91 & 1.21 & $9.34\%$  & $18.06\%$ & $9.55\%$ & $17.00\%$ \\
\midrule
  YOLOX-S+P4  & $640\times640$ & 9.69 & 60.14 & $21.71\%$ & $38.59\%$ & $18.26\%$ & $32.17\%$ \\
  YOLOX-Tiny+P4  & $448\times448$ &5.45 & 16.67 & $16.46\%$ & $29.97\%$ & $13.87\%$ & $24.97\%$ \\
  YOLOX-Nano+P4  & $448\times448$ &0.94 & 1.98 & $12.47\%$ & $22.01\%$ & $9.57\%$ & $18.49\%$\\
\midrule
  FasterX-S     & $640\times640$ & 5.19  & 19.20 & $22.37\%$ & $39.71\%$ & $19.29\%$ & $32.62\%$ \\
  FasterX-Tiny  & $448\times448$ & 2.93 & 5.39 & $17.26\%$ & $30.32\%$ & $14.65\%$ & $27.62\%$ \\
  FasterX-Nano  & $448\times448$ & 0.70 & 1.43 & $11.95\%$ & $21.02\%$ & $10.62\%$ & $19.39\%$ \\
  FasterX-S     & $832 \times838$ & 5.19  & 32.69 & $25.54\%$ & $46.69\%$ & $22.45\%$ & $37.13\%$ \\
  FasterX-Tiny  & $832 \times838$ & 2.93 & 18.57 & $22.89\%$ & $44.75\%$ & $19.45\%$ & $33.77\%$ \\
  FasterX-Nano  & $832 \times838$ & 0.70 & 4.92 & $16.44\%$ & $31.32\%$ & $14.31\%$ & $27.22\%$ \\
\bottomrule[1pt]
\end{tabular}
\label{table.comparison}
\end{table*}

\begin{table*}[!t]  \caption{Detection accuracy comparison of small objects, medium objects and large objects on Visdrone2021 dataset.}
 \label{table.dataset}
 \centering   
 \begin{tabular}{cccccccccc}
\toprule[1pt]
   Model & Size & Backbone &GFLOPs & $AP_{S}(val)$ & $AP_{M}(val)$ & $AP_{L}(val)$ & $AP_{S}(test)$ & $AP_{M}(test)$ & $AP_{L}(test)$\\
\midrule[1pt]
  YOLOX-S&      640 & 9.0  & 26.8& $12.4\%$  & $27.64\%$ &   $11.7\%$ &  $8.8\%$ &  $25.02\%$ &  $11.73\%$\\
  YOLOX-Tiny &  448 & 5.06 & 7.43 & $7.38\%$ & $23.07\%$ &  $11.82\%$ & $5.77\%$ & $20.75\%$ &$10.66\%$\\
  YOLOX-Nano &  448 & 0.91 & 1.21 & $5.26\%$ & $17.00\%$ &   $6.94\%$ & $3.66\%$ & $17.27\%$ & $6.18\%$\\
  \midrule
  FasterX-S&      640 & 5.19 & 19.20& $17.29\%$ & $29.73\%$ &  $12.23\%$ & $11.40\%$ & $27.20\%$ & $11.13\%$\\
  FasterX-Tiny &  448 & 2.93 & 5.39 & $10.63\%$ &  $26.36\%$ &  $14.31\%$ & $6.87\%$ & $23.81\%$ & $10.82\%$\\
  FasterX-Nano &  448 & 0.70 & 1.43& $7.73\%$ & $19.43\%$ &  $6.86\%$ & $5.17\%$ & $19.26\%$ & $6.32\%$\\
  \bottomrule[1pt]
\end{tabular}
\label{table.small object}
\end{table*}

\section{EXPERIMENTS}
\label{sec:7}
In this section, VisDrone2021 dataset and VOC2012 dataset are used to verify the validity of the proposed FasterX. We use NVIDIA Jetson NX and Jetson Nano as edge GPUs devices to evaluate the speed of FasterX, and report mAP (average of all 10 IOU thresholds, rank is [0.5:0.05:0.95]) and AP50 to evaluate the accuracy.
Meanwhile, we adopt FPS and latency ($ms$) to show the inference efficiency.
It should be noted that we conduct 10 rounds inference for each model to obtain averaged stable data, considering that the edge GPU devices may have relative errors on model inference. Specifically, each round inference contains 64 test images.

\subsection{Implementation Details}
\label{sec:8}
In this paper, all the experiments were completed in the deep learning framework -- pytorch version 1.8.0 and were trained and tested with the NVIDIA RTX TITAN GPU.
For the performance evaluation scheme on the Edge computing platform (Jetson NX and Jetson Nano), we select the commonly used ONNX as the intermediate expression, and further use TensorRT for serialization acceleration (coded storage format is FP16).
The Jetson NX and Nano platform environments are Jetpack 4.6 and Jetpack 4.5.1, respectively, and TensorRT versions are 8.0.1 and 7.1.3, respectively.
It is worth noting that Jetson NX is more powerful than Jetson Nano in terms of computational performance, Jetson Nano has a greater fluctuation in speed than NX under the same lightweight model conditions.

For a fair comparison, during the training, all models are trained from scratch and no pre-training weights are loaded.
For the training strategy, like YOLOX, we use the cosine annealing learning rate of warm-up. Considering the lack of extraction ability in the lightweight models, we only use mosaic enhancement to weak the data enhancement. Meanwhile, SGD optimizer is used in the training and momentum is set to 0.9.
It is worth noting that the input size of YOLOX-Tiny and YOLOX-Nano models is $416\times416$, and an odd number will occur after 32 times down-sampling, resulting in the fuzzy boundary of the scale in the prediction. Therefore, in the comparative experiment, the input size of the YOLOX-Tiny and YOLOX-Nano models is changed to $448\times448$, and the YOLOX-S model is still $640\times640$.

\begin{table*}[!t]
\caption{Ablation experiment on edge GPU devices.}
 \label{table.dataset}
 \centering   
 \begin{tabular}{c|c|c|c|c|c|c|c|c|c|c}
\toprule[1pt]
   Model(Size) &\thead{Feature\\Aggr.Path}& \thead{Head\\Deco.Mode}&\thead{Attention} &\thead{CIoU\\\&FL} &\thead{Aux} & \thead{Params\\(M)} &GFLOPs & \thead{FPS\\Nano\quad NX}& \thead{Latency(ms)\\Nano\quad NX} &\thead{Val\\mAP\quad AP50}\\
\midrule
  YOLOX-S&\cellcolor{blue!20}PANet &\cellcolor{blue!20}Conv& \cellcolor{blue!20}- &  \cellcolor{blue!20}- & \cellcolor{blue!20} -&\cellcolor{blue!20}9.0&\cellcolor{blue!20}26.8&\cellcolor{blue!20}19.04\quad68.02&\cellcolor{blue!20}52.52\quad14.70&\cellcolor{blue!20}$19.82\%$\quad$34.00\%$\\
  S+4 head&\cellcolor{pink!20}PANet &\cellcolor{pink!20}Conv& \cellcolor{pink!20}- &  \cellcolor{pink!20}- & \cellcolor{pink!20} - &\cellcolor{pink!20}9.69&\cellcolor{pink!20}60.14&\cellcolor{pink!20}4.19\quad36.68&\cellcolor{pink!20}238.22\quad27.26&\cellcolor{pink!20}$\textbf{21.71\%}$\quad\textbf{38.59\%}\\
  YOLOX-Tiny&\cellcolor{green!20}PANet &\cellcolor{green!20}Conv& \cellcolor{green!20}- & \cellcolor{green!20}- &\cellcolor{green!20}  -&\cellcolor{green!20}5.06&\cellcolor{green!20}7.43&\cellcolor{green!20}25.71\quad93.63&\cellcolor{green!20}38.89\quad10.68&\cellcolor{green!20}$14.37\%$\quad$25.33\%$\\
  Tiny+4 head&\cellcolor{blue!20}PANet&\cellcolor{blue!20}Conv& \cellcolor{blue!20}- &  \cellcolor{blue!20}- & \cellcolor{blue!20} -  &\cellcolor{blue!20}5.45&\cellcolor{blue!20}16.67&\cellcolor{blue!20}12.52\quad77.57&\cellcolor{blue!20}79.85\quad12.89&\cellcolor{blue!20}$16.46\%$\quad29.97\%\\
  YOLOX-Nano&\cellcolor{pink!20}PANet &\cellcolor{pink!20}DS Conv& \cellcolor{pink!20}- &  \cellcolor{pink!20}- &\cellcolor{pink!20}  -&\cellcolor{pink!20}0.91&\cellcolor{pink!20}1.21&\cellcolor{pink!20}\textbf{36.27}\quad\textbf{112.23}&\cellcolor{pink!20}\textbf{27.57}\quad\textbf{8.91}&\cellcolor{pink!20}$9.34\%$\quad$18.06\%$\\
  Nano+4 head&\cellcolor{green!20}PANet&\cellcolor{green!20}DS Conv& \cellcolor{green!20}- &  \cellcolor{green!20}- &  \cellcolor{green!20}- &\cellcolor{green!20}0.94&\cellcolor{green!20}1.98&\cellcolor{green!20}18.09\quad91.24&\cellcolor{green!20}55.25\quad10.96&\cellcolor{green!20}12.47\%\quad22.01\%\\
\midrule
\multirow{7}{*}{\thead{FasterX-S\\($640\times640$)}}
&\cellcolor{blue!20}PANet&\cellcolor{blue!20}DS Conv & \cellcolor{blue!20}- &  \cellcolor{blue!20}- &  \cellcolor{blue!20}- &\cellcolor{blue!20}7.61&\cellcolor{blue!20}24.88&\cellcolor{blue!20}11.10\quad52.00&\cellcolor{blue!20}90.09\quad19.23&\cellcolor{blue!20}$21.34\%$\quad$37.75\%$\\
&\cellcolor{pink!20}SlimFPN&\cellcolor{pink!20}DS Conv& \cellcolor{pink!20}- &  \cellcolor{pink!20}- & \cellcolor{pink!20} - &\cellcolor{pink!20}4.96&\cellcolor{pink!20}22.87&\cellcolor{pink!20}11.29\quad54.02&\cellcolor{pink!20}88.56\quad18.51&\cellcolor{pink!20}$20.68\%$\quad$36.52\%$\\
&\cellcolor{green!20}SlimFPN&\cellcolor{green!20}PixSF& \cellcolor{green!20}- &  \cellcolor{green!20}- &  \cellcolor{green!20}- &\cellcolor{green!20}7.25&\cellcolor{green!20}27.99&\cellcolor{green!20}6.47\quad42.88&\cellcolor{green!20}154.51\quad23.32&\cellcolor{green!20}21.93\%\quad$37.90\%$ \\
&\cellcolor{blue!20}SlimFPN&\cellcolor{blue!20}DS+PixSF& \cellcolor{blue!20}- & \cellcolor{blue!20} - & \cellcolor{blue!20} -  &\cellcolor{blue!20}5.19&\cellcolor{blue!20}19.20&\cellcolor{blue!20}\textbf{12.95\quad 59.88}&\cellcolor{blue!20}\textbf{77.21\quad 16.70}&\cellcolor{blue!20}$21.30\%$\quad $37.09\%$\\
&\cellcolor{pink!20}SlimFPN&\cellcolor{pink!20}DS+PixSF&\cellcolor{pink!20}$\surd$ & \cellcolor{pink!20} - & \cellcolor{pink!20} -  &\cellcolor{pink!20}5.19&\cellcolor{pink!20}19.20&\cellcolor{pink!20}11.68\quad 46.62&\cellcolor{pink!20}85.81\quad 21.45&\cellcolor{pink!20}$21.46\%$\quad $37.23\%$\\
&\cellcolor{green!20}SlimFPN&\cellcolor{green!20}DS+PixSF& \cellcolor{green!20}$\surd$ & \cellcolor{green!20} $\surd$ & \cellcolor{green!20} -  &\cellcolor{green!20}5.19&\cellcolor{green!20}19.20&\cellcolor{green!20}11.68\quad 46.62&\cellcolor{green!20}85.81\quad 21.45&\cellcolor{green!20}$21.84\%$\quad $38.62\%$\\
&\cellcolor{blue!20}SlimFPN&\cellcolor{blue!20}DS+PixSF& \cellcolor{blue!20}$\surd$ &  \cellcolor{blue!20}$\surd$ &  \cellcolor{blue!20}$\surd$ &\cellcolor{blue!20}5.19&\cellcolor{blue!20}19.20&\cellcolor{blue!20}11.68\quad 46.62&\cellcolor{blue!20}85.81\quad 21.45&\cellcolor{blue!20}$\textbf{22.37\%}$\quad $\textbf{39.71\%}$\\
\midrule
\multirow{7}{*}{\thead{FasterX-Tiny\\($448\times448$)}}
&\cellcolor{blue!20}PANet&\cellcolor{blue!20}DS Conv& \cellcolor{blue!20}- &  \cellcolor{blue!20}- &  \cellcolor{blue!20}- &\cellcolor{blue!20}4.29&\cellcolor{blue!20}6.99&\cellcolor{blue!20}14.67\quad72.15&\cellcolor{blue!20}68.12\quad13.86&\cellcolor{blue!20}$15.43\%$\quad $28.64\%$\\
&\cellcolor{pink!20}SlimFPN&\cellcolor{pink!20}DS Conv& \cellcolor{pink!20}- &  \cellcolor{pink!20}- &  \cellcolor{pink!20}-  &\cellcolor{pink!20}2.80&\cellcolor{pink!20}6.44&\cellcolor{pink!20}15.37\quad74.18&\cellcolor{pink!20}65.03\quad13.48&\cellcolor{pink!20}$15.12\%$\quad$27.99\%$\\
&\cellcolor{green!20}SlimFPN &\cellcolor{green!20}PixSF& \cellcolor{green!20}- &  \cellcolor{green!20}- &  \cellcolor{green!20}- &\cellcolor{green!20}4.09&\cellcolor{green!20}7.81&\cellcolor{green!20}17.01\quad\textbf{82.71}&\cellcolor{green!20}58.77\quad\textbf{12.09}&\cellcolor{green!20}16.60\%\quad28.90\%\\
&\cellcolor{blue!20}SlimFPN&\cellcolor{blue!20}DS+PixSF& \cellcolor{blue!20}- &  \cellcolor{blue!20}- &  \cellcolor{blue!20}- &\cellcolor{blue!20}2.93&\cellcolor{blue!20}5.39&\cellcolor{blue!20}\textbf{17.18}\quad81.83&\cellcolor{blue!20}\textbf{58.21}\quad12.22&\cellcolor{blue!20}$15.93\%$\quad$27.96\%$\\
&\cellcolor{pink!20}SlimFPN&\cellcolor{pink!20}DS+PixSF& \cellcolor{pink!20}$\surd$ &  \cellcolor{pink!20}- & \cellcolor{pink!20} - &\cellcolor{pink!20}2.93&\cellcolor{pink!20}5.39&\cellcolor{pink!20}16.24\quad76.33&\cellcolor{pink!20}61.54\quad13.10&\cellcolor{pink!20}$16.21\%$\quad$28.91\%$\\
&\cellcolor{green!20}SlimFPN&\cellcolor{green!20}DS+PixSF&\cellcolor{green!20} $\surd$ & \cellcolor{green!20} $\surd$ & \cellcolor{green!20} - &\cellcolor{green!20}2.93&\cellcolor{green!20}5.39&\cellcolor{green!20}16.24\quad76.33&\cellcolor{green!20}61.54\quad13.10&\cellcolor{green!20}$16.80\%$\quad$29.47\%$\\
&\cellcolor{blue!20}SlimFPN&\cellcolor{blue!20}DS+PixSF& \cellcolor{blue!20}$\surd$ &  \cellcolor{blue!20}$\surd$ &  \cellcolor{blue!20}$\surd$ &\cellcolor{blue!20}2.93&\cellcolor{blue!20}5.39&\cellcolor{blue!20}16.24\quad76.33&\cellcolor{blue!20}61.54\quad13.10&\cellcolor{blue!20}$\textbf{17.26\%}$\quad$\textbf{30.32\%}$\\
\midrule
\multirow{6}{*}{\thead{FasterX-Nano\\($448\times448$)}}
&\cellcolor{blue!20}SlimFPN&\cellcolor{blue!20}DS Conv& \cellcolor{blue!20}- &  \cellcolor{blue!20}- &  \cellcolor{blue!20}- &\cellcolor{blue!20}0.63&\cellcolor{blue!20}1.92&\cellcolor{blue!20}21.58\quad96.61&\cellcolor{blue!20}46.32\quad10.35&\cellcolor{blue!20}$10.73\%$\quad$21.13\%$\\
&\cellcolor{pink!20}SlimFPN &\cellcolor{pink!20}PixSF & \cellcolor{pink!20}- &  \cellcolor{pink!20}- &\cellcolor{pink!20}  -&\cellcolor{pink!20}1.21&\cellcolor{pink!20}2.49&\cellcolor{pink!20}21.93\quad\textbf{129.03}&\cellcolor{pink!20}45.58\quad\textbf{7.75}&\cellcolor{pink!20}$11.47\%$\quad$\textbf{21.91\%}$\\
&\cellcolor{green!20}SlimFPN&\cellcolor{green!20}DS+PixSF& \cellcolor{green!20}- & \cellcolor{green!20} - & \cellcolor{green!20} - &\cellcolor{green!20}0.70&\cellcolor{green!20}1.43&\cellcolor{green!20}\textbf{24.26}\quad102.98&\cellcolor{green!20}\textbf{41.22}\quad9.71&\cellcolor{green!20}$10.89\%$\quad$21.32\%$\\
&\cellcolor{blue!20}SlimFPN&\cellcolor{blue!20}DS+PixSF& \cellcolor{blue!20}$\surd$ &  \cellcolor{blue!20}- &  \cellcolor{blue!20}- &\cellcolor{blue!20}0.70&\cellcolor{blue!20}1.43&\cellcolor{blue!20}23.47\quad102.14&\cellcolor{blue!20}42.60\quad9.79&\cellcolor{blue!20}$11.14\%$\quad$20.70\%$\\
&\cellcolor{pink!20}SlimFPN&\cellcolor{pink!20}DS+PixSF& \cellcolor{pink!20}$\surd$ &  \cellcolor{pink!20}$\surd$ &  \cellcolor{pink!20}- &\cellcolor{pink!20}0.70&\cellcolor{pink!20}1.43&\cellcolor{pink!20}23.47\quad102.14&\cellcolor{pink!20}42.60\quad9.79&\cellcolor{pink!20}$11.33\%$\quad$20.42\%$\\
&\cellcolor{green!20}SlimFPN&\cellcolor{green!20}DS+PixSF& \cellcolor{green!20}$\surd$ &  \cellcolor{green!20}$\surd$ &  \cellcolor{green!20}$\surd$ &\cellcolor{green!20}0.70&\cellcolor{green!20}1.43&\cellcolor{green!20}23.47\quad102.14&\cellcolor{green!20}42.60\quad9.79&\cellcolor{green!20}$\textbf{11.95\%}$\quad$21.02\%$\\
\bottomrule[1pt]
\end{tabular}
 \label{table.ablation}
\end{table*}

\subsection{Performance Comparison}
In order to demonstrate the superior performance of FasterX, we conducted a series of experiments on Visdrone2021 dataset with other state-of-the-art one-stage detectors, such as SlimYOLOv3, YOLOV5, and YOLOX.
Experimental results are shown in Table \ref{table.comparison}. It is worth noting that the results of these compared methods are referenced from the corresponding original papers.

From Table~\ref{table.comparison}, it can be observed that FasterX-Tiny reduces $43.1\%$ parameters and $29.4\%$ GFLOPs, and achieves $1.69\%$ higher mAp and $8.65\%$ higher AP50 than SlimYOLOv3-SPP3-95. FasterX-S achieves $4.34\%$ higher mAp and $10.59\%$ higher AP50 than SlimYOLOv3-SPP3-95 \cite{20}, and only increase $1.8\%$ parameters.
Obviously, YOLOX achieves higher accuracy than YOLOv5 in all model and has lower parameters and GFLOPs in S and Nano. In particular, with the addition of P4 head, 4-head YOLOX series have a greater improvement in detection accuracy. However, parameters and GFLOPs increase significantly with the added P4 head of YOLOX series.
As shown in Table \ref{table.comparison} and Fig.~\ref{Fig.system}, FasterX series (S, Tiny and Nano) provide not only a much better mAp and AP50 but also a lower parameters and GFLOPs that YOLOv5 series, YOLOX series and 4-head YOLOX series, thanks to the better feature expression of the lightweight PixSF head and online distillation mechanism.
Experimental results demonstrate that the proposed FasterX with lightweight PixSF-head might be more powerful and effective than a 4-head YOLOX.

To illustrate the efficiency of FasterX on detecting small objects, we evaluate detection accuracy on the validation set and test set of Visdrone2021 dataset. Experimental results are shown shown in Table~\ref{table.small object}. Obviously, FasterX series are much better than YOLOX series in detection accuracy.

From Table~\ref{table.small object}, it can be observed that FasterX series achieve a greater accuracy improvement on small objects. In contrast, the accuracy improvement is not obvious on large objects. For example, the detection accuracy of FasterX-S on small objects, medium objects and large objects are increased by 4.89\%, 2.09\% and 0.53\%, respectively. Such results show that the efficiency of the lightweight PixSF head for small objects.

\begin{table*}[!t] \caption{Comparison of the different Head and Feature Aggregation with edge GPU on VOC2012 dataset.}
 \label{table.dataset}
 \centering   
 \begin{tabular}{c|c|c|c|c|c}
\toprule[1pt]
   Model &\thead{Feature\\Aggr.Path} & \thead{Head\\Deco.Mode}& \thead{Params\\(M)} &GFLOPs
 &\thead{Val\\mAP\quad AP50}\\
\hline\hline
\multirow{8}{*}{YOLOX-S}
  &  PANet &  Conv& 8.95 & 26.68 & $42.61\%$ \quad $64.52\%$ \\
  &  PANet & DS Conv     & 7.39 &17.96 & $39.12\%$ \quad $61.82\%$ \\
  &  PANet & PixSF       &7.00  &16.54 & $42.53\%$ \quad $63.81\%$ \\
  &  PANet & DS+PixSF    & 5.44 & 14.37& $40.20\%$ \quad $63.73\%$ \\
  &  SlimFPN & Conv& 6.34 & 23.04& $38.97\%$ \quad $63.50\%$ \\
  &  SlimFPN & DS Conv     & 4.79 &14.32 & $38.42\%$ \quad $63.43\%$ \\
  &  SlimFPN & PixSF       &6.52  &15.59 & $39.42\%$ \quad $63.44\%$ \\
  &  SlimFPN & DS+PixSF    &4.96  &13.41 & $39.00\%$ \quad $63.13\%$ \\

\hline\hline
\multirow{8}{*}{YOLOX-Tiny}
   &   PANet & Conv& 5.04 &7.43 & $34.92\%$ \quad $58.28\%$ \\
   &   PANet & DS Conv     &4.17  &5.04 & $34.67\%$ \quad $57.13\%$ \\
   &   PANet & PixSF       &3.95  &4.64 & $35.43\%$ \quad $57.91\%$ \\
   &   PANet & DS+PixSF    &3.07  &4.04 & $34.88\%$ \quad $57.60\%$ \\
   &   SlimFPN& Conv& 3.57 &6.43 &$33.53 \%$ \quad $56.87\%$ \\
   &   SlimFPN& DS Conv     & 2.70 &4.04 &$32.91 \%$ \quad $56.42\%$ \\
   &   SlimFPN& PixSF       &3.66  &4.36 &$34.19 \%$ \quad $57.75\%$ \\
   &   SlimFPN& DS+PixSF    &2.78  &3.76 &$33.44 \%$ \quad $56.83\%$ \\
\hline\hline
\multirow{6}{*}{YOLOX-Nano}
   &   PANet   & DS Conv &0.90  &1.21 & $25.30\%$  \quad $47.31\%$ \\
   &   PANet   & PixSF   &1.06  & 1.19& $27.37\%$  \quad $48.68\%$ \\
   &   PANet   & DS+PixSF &0.68  &0.92 & $26.78\%$  \quad $48.03\%$ \\
   &   SlimFPN & DS Conv &0.60  &1.03 & $24.68\%$  \quad $47.21\%$ \\
   &   SlimFPN & PixSF   &1.02  &1.15  & $26.42\%$  \quad $48.49\%$ \\
   &   SlimFPN & DS+PixSF&0.64  &0.89 & $25.97\%$  \quad $47.38\%$ \\
\bottomrule[1pt]
\end{tabular}
\label{VOC2012}
\end{table*}

\subsection{Ablation Study}
\label{sec:9}
In order to reveal the impact of different tricks (such as Slim-FPN, PixSF-head, attention mechanism, SimOTA and online distillation) in FasterX on accuracy and computation complexity and latency, we conducted ablation study to prove the effectiveness of the proposed method. The results of the ablation experiments are shown in Table~\ref{table.ablation}. From Table~\ref{table.ablation}, it can be observed that the proposed FasterX series can not only increase mAP accuracy and AP50 accuracy, but also improve the inference speed and reduce the size of the network.
Next, we will discuss the result of ablation experiment in detail.

\textbf{PixSF head and SlimFPN}:
We first explore the effect of 4-head structure for YOLOX. Taking 4-head YOLOX-S as an example, as shown in Table~\ref{table.ablation} (row 2), the mAP and AP50 are increased significantly by 1.89\% and 4.59\% respectively, which verifies the efficiency of the 4-head structure. However, adding a detection head makes the GFLOPs change from 26.8 to 60.14, and latency from 52.52ms to 238.22ms on Jetson Nano, 14.70ms to 27.26ms on Jetson NX. In order to alleviate this situation, replacing the convolution operator with DS Conv in the head layer is a feasible design for lightweight model\cite{21,13}.
As shown in Table~\ref{table.ablation} (FasterX-S part, row 1 and FasterX-Tiny part, row 1), replacing the general convolution with the DS Conv, although the mAP and AP50 decrease a little bit, the latency has decreased significantly. Experimental results show the efficiency of the DS Conv operator.

To demonstrate the efficient performance of our PixSF head approach, we compare the general convolution operator, DW Conv operator as well as our PixSF operator. To show the flexible embeddedness of PixSF head, we also embed DW Conv into PixSF head to design a more lightweight head layer. As shown in Table~\ref{table.ablation} (FasterX-S part, row 2 and 4, FasterX-Tiny part, row 2 and 4, FasterX-Nano part, row 1 and 3), it can be observed that our DW+PixSF method show better performance than general convolution operator and DW operator in latency and detection accuracy. Taking FasterX-S as an example, compared to 4-head structure with DW operator, not only the mAP and AP50 are decreased by 0.62\% and 0.57\% respectively, but also the latency is decreased by over $12.8\%$ and $9.8\%$ on Jetson Nano and NX, respectively.
Such results illustrate that the proposed PixSF head can not only efficiency on detection accuracy but also improve the inference speed.

Next, we illustrate the trade-off between accuracy and speed in the feature aggregation part. As shown in Tabel \ref{table.ablation} (FasterX-S part, row 2, FasterX-Tiny part, row 2 and FasterX-Nano part, row 1), taking FasterX-S as an example, compared with the PANet (FasterX-S part, row 1), although the mAP and AP50 of SlimFPN method are decreased by 0.66\% and 1.23\% respectively, the parameters decrease from 7.61 to 4.96, and GFLOPs from 24.88 to 22.87. Such results imply that SlimFPN not only reduce the size of of the network but also can keep the detection accuracy. This is because the top-down structure can ensure that the deep semantic information is transferred to shallow feature maps, which can provide semantic support for small objects.

To further validate the generality of the PixSF head, we performed experiments on VOC2012 dataset with a higher number of larger object than UAV dataset. The experimental results in Table \ref{VOC2012} show that the combination of DS Conv and PixSF head can further achieve the trade off between the model capacity and accuracy.

\textbf{Attention mechanism:}
To enhance the decoupling performance of Head layer of object detection, we adopt CBAM to improve the feature representation of Head layer. As shown in Tabel \ref{table.ablation} (FasterX-S part, row 5, FasterX-Tiny part, row 5 and FasterX-Nano part, row 4), it can be seen that CBAM present positive impacts on the accuracy. Because it not only monitors the channel, but also extracts the interest area by using the spatial probability map.

\textbf{Improved SimOTA:}
To verify the effectiveness of the proposed dynamic label assignment strategy, we compare the improved SimOTA with the basic label assignment mechanism. As shown in TABLE~\ref{table.ablation}, we replace the original SimOTA with the improved SimOTA.
The experimental results show that the improved SimOTA can achieve good results in all three models. Taking FasterX-S as an example (FasterX-S part, row 6), the mAP and AP50 are increased by 0.42\% and 1.39\% respectively, without additional computing resource.

\textbf{Auxiliary Head:}
In addition, to illustrate the efficiency of the auxiliary head for online distillation, we explore the effect of the auxiliary head for FasteX. In order to improve the extraction ability of auxiliary head, we use the YOLOX-X head with a large number of parameters for training supervise. In the process of training, we adopt the network preheating strategy. First, the PixSF head and auxiliary head are trained jointly for 50 epochs. Then, we use label results of the auxiliary head to guide the PixSF head. It can be observed from TABLE~\ref{table.ablation} that mAP and AP50 have shown significant improvement after using the auxiliary head for online distillation.

\begin{table*}[!t]\caption{Evaluation results of Backbone accuracy and Latency on VisDrone2021 dataset.}
 \centering   
\label{table.dataset}
 \begin{tabular}{c|c|c|c|c|c|c|c}
\toprule[2pt]
   Model(Size)  & \thead{Params\\(M)} &GFLOPs &Backbone & AP50(val) & AP50(test) & \thead{FPS\\Nano\quad NX } & \thead{Latency\\Nano\quad NX} \\
\midrule[1pt]
\multirow{12}{*}{\thead{YOLOX-S\\+P4 ($640\times640$)}}
      & 7.25 & 27.99& CSPDarknet53 &$37.92\%$& $31.63\%$ &6.47\quad 42.88 &154.51ms\quad 23.32ms\\
      & 4.79 & 21.66& MblNetV2(width 1)&$35.87\%$& $29.55\%$ &4.82\quad44.66 & 207.09ms\quad 22.39ms\\
       &4.01&20.28& MblNetV2(width 0.75) &$35.43\%$&$28.97\%$ &5.02\quad45.87 & 198.81ms\quad 21.80ms\\
      & 3.44& 18.85& MblNetV2(width 0.5) &$34.61\%$&$28.22\%$&5.70\quad \textbf{49.97} & 175.29ms\quad \textbf{20.01ms}\\
     &7.19 &21.03& GhostNet(width 1.3) &$36.12\%$& $30.12\%$ &4.68\quad 34.28 & 213.25ms\quad 29.17ms \\
     &5.48& 19.43& GhostNet(width 1) &$36.38\%$& $29.66\%$ &5.26\quad 38.13 & 189.90ms\quad 26.22ms \\
    &3.59& 17.54& GhostNet(width 0.5) &$31.40\%$& $25.30\%$ &\textbf{6.61}\quad 46.77 & \textbf{151.16ms}\quad 21.38ms\\
     & 5.94 & 23.06& Efficientnet-Lite0 &$38.98\%$& $32.52\%$ &4.55\quad 44.80 & 219.53ms\quad 22.32ms \\
     & 6.71 & 25.04& Efficientnet-Lite1 &$38.56\%$& $31.98\%$ &3.76\quad 39.41 & 265.85ms\quad 25.37ms \\
       & 7.35& 26.25& Efficientnet-Lite2  &$39.11\%$& $33.31\%$ &3.65\quad 38.18 & 273.54ms\quad 26.19ms\\
       &9.42& 30.84& Efficientnet-Lite3  &$39.32\%$& $33.82\%$ &3.05\quad 33.20 & 326.83ms\quad 30.12ms\\
    &14.16 &39.02 & Efficientnet-Lite4 &$\textbf{40.16\%}$& $\textbf{34.11\%}$ &2.35\quad 25.89 & 424.98ms\quad 38.62ms\\
\midrule[1pt]
\multirow{12}{*}{\thead{YOLOX-Tiny\\+P4 ($448\times448$)}}
& 4.09 & 7.81 & CSPDarknet53 &$28.90\%$& $25.19\%$ & \textbf{17.01}\quad \textbf{82.71} &\textbf{58.77ms}\quad \textbf{12.09ms}\\
& 3.51 & 7.03 & MblNetV2(width 1)&$28.10\%$& $23.44\%$ & 11.53\quad 81.69 & 86.71ms\quad 12.24ms\\
 &2.73&6.36 & MblNetV2(width 0.75) &$28.32\%$&$24.13\%$ & 14.35\quad 81.36&79.64ms\quad 12.29ms\\
 & 2.16 & 5.67 & MblNetV2(width 0.5)&$27.39\%$&$22.85\%$& 14.52\quad 81.30 &68.83ms\quad 12.30ms\\
 & 5.91 & 6.72 & GhostNet(width 1.3) &$28.08\%$&  $24.44\%$ & 10.46\quad 66.40 & 95.55ms\quad 15.06ms \\
 &4.20&5.94& GhostNet(width 1) &$27.39\%$& $23.56\%$ & 11.89\quad 72.72 & 84.05ms\quad 13.75ms \\
 &2.32&5.03& GhostNet(width 0.5) &$24.24\%$& $20.66\%$ & 15.64\quad 68.44 & 63.93ms\quad 14.61ms \\
& 4.66  & 7.72  & Efficientnet-Lite0 & $28.32\%$ &  $24.33\%$ & 10.16\quad 78.18 & 98.40ms\quad 12.79ms \\
& 5.42  & 8.69  & Efficientnet-Lite1 & $28.36\%$ &  $24.27\%$ & 8.37\quad 73.74 & 119.43ms\quad13.56ms  \\
& 6.06  & 9.28  & Efficientnet-Lite2 & $29.21\%$ &  $24.90\%$ & 7.83\quad 73.47 & 127.59ms\quad 13.61ms \\
& 8.13  & 11.52 & Efficientnet-Lite3 & $30.60\%$  &  $24.81\%$ & 6.56\quad 64.26 & 152.29ms\quad 15.56ms\\
& 12.87 & 15.53 & Efficientnet-Lite4 & $\textbf{31.24\%}$ &  $\textbf{26.79\%}$ & 4.92\quad 51.38 & 203.03ms\quad 19.46ms\\
\midrule[1pt]
\multirow{12}{*}{\thead{YOLOX-Nano\\+P4 ($448\times448$)}}
& 0.69 & 1.43 & CSPDarknet53       &$20.32\%$& $18.45\%$ & \textbf{24.26}\quad\textbf{102.98} & \textbf{41.22ms}\quad \textbf{9.71ms}\\
& 2.05 & 3.13 & MblNetV2(width 1)&$24.58\%$& $21.02\%$ & 13.31\quad 91.40  & 75.08ms\quad 10.94ms\\
& 1.27&2.47 & MblNetV2(width 0.75) &$23.78\%$&$20.77\%$ & 14.55\quad 93.10 & 68.69ms\quad 10.74ms\\
& 0.70 & 1.78 & MblNetV2(width 0.5)&$22.10\%$&$20.13\%$ & 18.04\quad 93.80  & 55.43ms\quad 10.66ms\\
& 4.45 &2.81 & GhostNet(width 1.3) &$24.74\%$& $21.55\%$ & 12.07\quad 67.15& 82.81ms\quad 14.89ms\\
& 2.74& 2.04  & GhostNet(width 1) &$23.59\%$& $20.00\%$ & 14.45\quad 67.38 & 69.19ms\quad 14.84ms\\
& 0.87&1.14 & GhostNet(width 0.5) &$19.39\%$& $17.33\%$ & 19.90\quad 84.74 & 50.23ms\quad 11.80ms\\
& 3.19 & 3.82 & Efficientnet-Lite0 &$24.98\%$& $22.71\%$ & 11.74\quad 77.57  & 85.16ms\quad 12.89ms\\
& 3.96 &4.79& Efficientnet-Lite1   &$24.30\%$& $22.85\%$ & 9.59\quad 73.26   & 104.24ms\quad 13.65ms\\
& 4.6& 5.37 & Efficientnet-Lite2   &$25.10\%$& $23.12\%$ & 8.75\quad 69.10   & 114.19ms\quad 14.47ms\\
& 6.66&7.62 & Efficientnet-Lite3   &$25.35\%$& $23.33\%$ & 7.08\quad 64.59   & 141.20ms\quad 15.48ms\\
& 11.40 &11.62 & Efficientnet-Lite4 &$\textbf{25.84\%}$& $\textbf{23.93\%}$ & 5.19\quad 59.80  & 192.46ms\quad 16.72ms\\
\bottomrule[2pt]
\end{tabular}
\label{backbone}
\end{table*}

\textbf{Backbone:}
In this paper, we employ CSPDarknet53 as the backbone of FasterX. Instead of theoretically modeling the relation between backbones and inference speed, we directly report the FPS and latency of the current popular lightweight backbones (such as MobileNetV2, GhostNet and Efficientnet-Lite) under the 4-head structure on Jetson devices. As shown in Table \ref{backbone}, it is observed that Efficientnet-Lite4 achieves the best detection accuracy. At the same time, the latency is highest among all backbones. CSPDarknet53 backbone is able to operate with a higher detection accuracy without sacrificing more computing time. Hence, a balance is achieved in the CSPDarknet53 backbone between detection accuracy and inference speed.

\section{Conclusion}
\label{sec:1}
In this paper, we propose a novel lightweight object detector FasterX for UAV detection on edge GPU devices. The most important part of the proposed FasterX is the lightweight PixSF head, where the position encode-decoder is designed to improve the detection accuracy of small objects by using the position embedding in the head layer. Meanwhile, PixSF head can be further embedded in the depthwise separable convolution to obtain a lighter head. Furthermore, SlimFPN is developed to boost the inference speed. In additional, attention mechanism, label assignment strategy and auxiliary head methods are presented to further improve the object detection accuracy of the model.
The efficacy of the FasterX is validated experimentally on VisDrone2021 dataset, which has a trade-off between accuracy and speed. Extensive experiments demonstrate that our method can obtain better performance than state-of-the-art detection methods in terms of accuracy and speed.

Considering the disadvantages of dense detector in UAV, a large number of target candidates lead to redundant calculation, which often requires NMS to remove redundancy. In future work, we will focus the possible paradigm of lightweight design with sparse technology for dense detector.

\ifCLASSOPTIONcaptionsoff
  \newpage
\fi
{
\footnotesize
\bibliographystyle{IEEEtran}
\bibliography{zhou}

\begin{thebibliography}{10}
\providecommand{\url}[1]{#1}
\csname url@samestyle\endcsname
\providecommand{\newblock}{\relax}
\providecommand{\bibinfo}[2]{#2}
\providecommand{\BIBentrySTDinterwordspacing}{\spaceskip=0pt\relax}
\providecommand{\BIBentryALTinterwordstretchfactor}{4}
\providecommand{\BIBentryALTinterwordspacing}{\spaceskip=\fontdimen2\font plus
\BIBentryALTinterwordstretchfactor\fontdimen3\font minus
  \fontdimen4\font\relax}
\providecommand{\BIBforeignlanguage}[2]{{%
\expandafter\ifx\csname l@#1\endcsname\relax
\typeout{** WARNING: IEEEtran.bst: No hyphenation pattern has been}%
\typeout{** loaded for the language `#1'. Using the pattern for}%
\typeout{** the default language instead.}%
\else
\language=\csname l@#1\endcsname
\fi
#2}}
\providecommand{\BIBdecl}{\relax}
\BIBdecl

\bibitem{IOT1}
K.~Li, W.~Ni, E.~Tovar, and M.~Guizani, ``Joint flight cruise control and data
  collection in uav-aided internet of things: An onboard deep reinforcement
  learning approach,'' \emph{IEEE Internet of Things Journal}, vol.~8, no.~12,
  pp. 9787--9799, 2021.

\bibitem{IOT2}
X.~Huang, X.~Yang, Q.~Chen, and J.~Zhang, ``Task offloading optimization for
  uav-assisted fog-enabled internet of things networks,'' \emph{IEEE Internet
  of Things Journal}, vol.~PP, no.~99, pp. 1--1, 2021.

\bibitem{1}
N.~Audebert, B.~{Le Saux}, and S.~Lefèvre, ``Beyond rgb: Very high resolution
  urban remote sensing with multimodal deep networks,'' \emph{ISPRS Journal of
  Photogrammetry and Remote Sensing}, vol. 140, pp. 20--32, 2018.

\bibitem{2}
Z.~Shao, C.~Li, D.~Li, O.~Altan, L.~Zhang, and L.~Ding, ``An accurate matching
  method for projecting vector data into surveillance video to monitor and
  protect cultivated land,'' \emph{ISPRS Int. J. Geo Inf.}, vol.~9, p. 448,
  2020.

\bibitem{3}
``Detecting mammals in uav images: Best practices to address a substantially
  imbalanced dataset with deep learning,'' \emph{Remote Sensing of
  Environment}, vol. 216, pp. 139--153, 2018.

\bibitem{4}
B.~Kellenberger, M.~Volpi, and D.~Tuia, ``Fast animal detection in uav images
  using convolutional neural networks,'' in \emph{2017 IEEE International
  Geoscience and Remote Sensing Symposium (IGARSS)}, 2017, pp. 866--869.

\bibitem{5}
R.~Girshick, ``Fast r-cnn,'' in \emph{2015 IEEE International Conference on
  Computer Vision (ICCV)}, 2015, pp. 1440--1448.

\bibitem{6}
S.~Ren, K.~He, R.~Girshick, and J.~Sun, ``Faster r-cnn: Towards real-time
  object detection with region proposal networks,'' \emph{IEEE Transactions on
  Pattern Analysis and Machine Intelligence}, vol.~39, no.~6, pp. 1137--1149,
  2017.

\bibitem{7}
T.-Y. Lin, P.~Dollár, R.~Girshick, K.~He, B.~Hariharan, and S.~Belongie,
  ``Feature pyramid networks for object detection,'' in \emph{2017 IEEE
  Conference on Computer Vision and Pattern Recognition (CVPR)}, 2017, pp.
  936--944.

\bibitem{8}
X.~Long, K.~Deng, G.~Wang, Y.~Zhang, Q.~Dang, Y.~Gao, H.~Shen, J.~Ren, S.~Han,
  E.~Ding, and S.~Wen, ``Pp-yolo: An effective and efficient implementation of
  object detector,'' \emph{ArXiv}, vol. abs/2007.12099, 2020.

\bibitem{9}
P.~Ganesh, Y.~Chen, Y.~Yang, D.~Chen, and M.~Winslett, ``Yolo-ret: Towards high
  accuracy real-time object detection on edge gpus,'' in \emph{2022 IEEE/CVF
  Winter Conference on Applications of Computer Vision (WACV)}, 2022, pp.
  1311--1321.

\bibitem{10}
C.-Y. Wang, A.~Bochkovskiy, and H.-Y.~M. Liao, ``Scaled-yolov4: Scaling cross
  stage partial network,'' in \emph{2021 IEEE/CVF Conference on Computer Vision
  and Pattern Recognition (CVPR)}, 2021, pp. 13\,024--13\,033.

\bibitem{11}
Z.~Tian, C.~Shen, H.~Chen, and T.~He, ``Fcos: Fully convolutional one-stage
  object detection,'' in \emph{2019 IEEE/CVF International Conference on
  Computer Vision (ICCV)}, 2019, pp. 9626--9635.

\bibitem{12}
C.~Zhu, Y.~He, and M.~Savvides, ``Feature selective anchor-free module for
  single-shot object detection,'' in \emph{2019 IEEE/CVF Conference on Computer
  Vision and Pattern Recognition (CVPR)}, 2019, pp. 840--849.

\bibitem{13}
RangiLyu, ``Nanodet-plus: Super fast and high accuracy lightweight anchor-free
  object detection model.'' \url{https://github.com/RangiLyu/nanodet}, 2021.

\bibitem{14}
Z.~Ge, S.~Liu, F.~Wang, Z.~Li, and J.~Sun, ``Yolox: Exceeding yolo series in
  2021,'' \emph{arXiv preprint arXiv:2107.08430}, 2021.

\bibitem{15}
T.-Y. Lin, M.~Maire, S.~Belongie, J.~Hays, P.~Perona, D.~Ramanan,
  P.~Doll{\'a}r, and C.~L. Zitnick, ``Microsoft coco: Common objects in
  context,'' \emph{european conference on computer vision}, 2014.

\bibitem{16}
M.~Everingham, L.~V. Gool, C.~K.~I. Williams, J.~Winn, and A.~Zisserman, ``The
  pascal visual object classes (voc) challenge,'' \emph{International Journal
  of Computer Vision}, vol.~88, pp. 303--308, September 2009.

\bibitem{18}
X.~Zhu, S.~Lyu, X.~Wang, and Q.~Zhao, ``Tph-yolov5: Improved yolov5 based on
  transformer prediction head for object detection on drone-captured
  scenarios,'' in \emph{2021 IEEE/CVF International Conference on Computer
  Vision Workshops (ICCVW)}, 2021, pp. 2778--2788.

\bibitem{17}
P.~Zhu, L.~Wen, D.~Du, X.~Bian, H.~Fan, Q.~Hu, and H.~Ling, ``Detection and
  tracking meet drones challenge,'' \emph{IEEE Transactions on Pattern Analysis
  and Machine Intelligence}, pp. 1--1, 2021.

\bibitem{19}
J.~Deng, Z.~Shi, and C.~Zhuo, ``Energy-efficient real-time uav object detection
  on embedded platforms,'' \emph{IEEE Transactions on Computer-Aided Design of
  Integrated Circuits and Systems}, vol.~39, no.~10, pp. 3123--3127, 2020.

\bibitem{20}
P.~Zhang, Y.~Zhong, and X.~Li, ``Slimyolov3: Narrower, faster and better for
  real-time uav applications,'' \emph{2019 IEEE/CVF International Conference on
  Computer Vision Workshop (ICCVW)}, pp. 37--45, 2019.

\bibitem{21}
G.~Yu, Q.~Chang, W.~Lv, C.~Xu, C.~Cui, W.~Ji, Q.~Dang, K.~Deng, G.~Wang, Y.~Du,
  B.~Lai, Q.~Liu, X.~Hu, D.~Yu, and Y.~Ma, ``Pp-picodet: A better real-time
  object detector on mobile devices,'' \emph{ArXiv}, 2021.

\bibitem{22}
W.~Shi, J.~Caballero, F.~Huszár, J.~Totz, A.~P. Aitken, R.~Bishop,
  D.~Rueckert, and Z.~Wang, ``Real-time single image and video super-resolution
  using an efficient sub-pixel convolutional neural network,'' in \emph{2016
  IEEE Conference on Computer Vision and Pattern Recognition (CVPR)}, 2016, pp.
  1874--1883.

\bibitem{24}
K.~Han, Y.~Wang, Q.~Tian, J.~Guo, C.~Xu, and C.~Xu, ``Ghostnet: More features
  from cheap operations,'' in \emph{2020 IEEE/CVF Conference on Computer Vision
  and Pattern Recognition (CVPR)}, 2020, pp. 1577--1586.

\bibitem{42}
Z.~Ge, S.~Liu, Z.~Li, O.~Yoshie, and J.~Sun, ``Ota: Optimal transport
  assignment for object detection,'' in \emph{2021 IEEE/CVF Conference on
  Computer Vision and Pattern Recognition (CVPR)}, 2021, pp. 303--312.

\bibitem{25}
T.-Y. Lin, P.~Goyal, R.~Girshick, K.~He, and P.~Dollár, ``Focal loss for dense
  object detection,'' in \emph{2017 IEEE International Conference on Computer
  Vision (ICCV)}, 2017, pp. 2999--3007.

\bibitem{26}
Z.~Zheng, P.~Wang, W.~Liu, J.~Li, R.~Ye, and D.~Ren, ``Distance-iou loss:
  Faster and better learning for bounding box regression,'' in \emph{AAAI},
  2020.

\bibitem{27}
K.~Simonyan and A.~Zisserman, ``Very deep convolutional networks for
  large-scale image recognition,'' \emph{arXiv preprint arXiv:1409.1556}, 2014.

\bibitem{28}
K.~He, X.~Zhang, S.~Ren, and J.~Sun, ``Deep residual learning for image
  recognition,'' in \emph{2016 IEEE Conference on Computer Vision and Pattern
  Recognition (CVPR)}, 2016, pp. 770--778.

\bibitem{29}
G.~Huang, Z.~Liu, L.~Van Der~Maaten, and K.~Q. Weinberger, ``Densely connected
  convolutional networks,'' in \emph{2017 IEEE Conference on Computer Vision
  and Pattern Recognition (CVPR)}, 2017, pp. 2261--2269.

\bibitem{30}
Z.~Liu, Y.~Lin, Y.~Cao, H.~Hu, Y.~Wei, Z.~Zhang, S.~Lin, and B.~Guo, ``Swin
  transformer: Hierarchical vision transformer using shifted windows,'' in
  \emph{2021 IEEE/CVF International Conference on Computer Vision (ICCV)},
  2021, pp. 9992--10\,002.

\bibitem{31}
A.~G. Howard, M.~Zhu, B.~Chen, D.~Kalenichenko, W.~Wang, T.~Weyand,
  M.~Andreetto, and H.~Adam, ``Mobilenets: Efficient convolutional neural
  networks for mobile vision applications,'' \emph{ArXiv}, vol. abs/1704.04861,
  2017.

\bibitem{32}
X.~Zhang, X.~Zhou, M.~Lin, and J.~Sun, ``Shufflenet: An extremely efficient
  convolutional neural network for mobile devices,'' in \emph{2018 IEEE/CVF
  Conference on Computer Vision and Pattern Recognition}, 2018, pp. 6848--6856.

\bibitem{41}
X.~Ding, X.~Zhang, N.~Ma, J.~Han, G.~Ding, and J.~Sun, ``Repvgg: Making
  vgg-style convnets great again,'' in \emph{2021 IEEE/CVF Conference on
  Computer Vision and Pattern Recognition (CVPR)}, 2021, pp. 13\,728--13\,737.

\bibitem{33}
S.~Liu, L.~Qi, H.~Qin, J.~Shi, and J.~Jia, ``Path aggregation network for
  instance segmentation,'' in \emph{2018 IEEE/CVF Conference on Computer Vision
  and Pattern Recognition}, 2018, pp. 8759--8768.

\bibitem{34}
G.~Ghiasi, T.-Y. Lin, and Q.~V. Le, ``Nas-fpn: Learning scalable feature
  pyramid architecture for object detection,'' in \emph{2019 IEEE/CVF
  Conference on Computer Vision and Pattern Recognition (CVPR)}, 2019, pp.
  7029--7038.

\bibitem{35}
M.~Tan, R.~Pang, and Q.~Le, ``Efficientdet: Scalable and efficient object
  detection,'' 06 2020, pp. 10\,778--10\,787.

\bibitem{36}
S.~Liu, D.~Huang, and Y.~Wang, ``Learning spatial fusion for single-shot object
  detection,'' \emph{ArXiv}, vol. abs/1911.09516, 2019.

\bibitem{37}
Q.~Zhao, T.~Sheng, Y.~Wang, Z.~Tang, Y.~Chen, L.~Cai, and H.~Ling, ``M2det: A
  single-shot object detector based on multi-level feature pyramid network,''
  \emph{Proceedings of the AAAI Conference on Artificial Intelligence},
  vol.~33, pp. 9259--9266, 07 2019.

\bibitem{38}
Q.~Chen, Y.~Wang, T.~Yang, X.~Zhang, J.~Cheng, and J.~Sun, ``You only look
  one-level feature,'' in \emph{2021 IEEE/CVF Conference on Computer Vision and
  Pattern Recognition (CVPR)}, 2021, pp. 13\,034--13\,043.

\bibitem{39}
S.~Xu, X.~Wang, W.~Lv, Q.~Chang, C.~Cui, K.~Deng, G.~Wang, Q.~Dang, S.~Wei,
  Y.~Du, and B.~Lai, ``Pp-yoloe: An evolved version of yolo,'' \emph{ArXiv},
  vol. abs/2203.16250, 2022.

\bibitem{40}
C.-Y. Wang, A.~Bochkovskiy, and H.-Y.~M. Liao, ``{YOLOv7}: Trainable
  bag-of-freebies sets new state-of-the-art for real-time object detectors,''
  \emph{arXiv preprint arXiv:2207.02696}, 2022.

\bibitem{43}
Y.~Wu, Y.~Chen, L.~Yuan, Z.~Liu, L.~Wang, H.~Li, and Y.~Fu, ``Rethinking
  classification and localization for object detection,'' in \emph{2020
  IEEE/CVF Conference on Computer Vision and Pattern Recognition (CVPR)}, 2020,
  pp. 10\,183--10\,192.

\bibitem{44}
G.~Song, Y.~Liu, and X.~Wang, ``Revisiting the sibling head in object
  detector,'' in \emph{2020 IEEE/CVF Conference on Computer Vision and Pattern
  Recognition (CVPR)}, 2020, pp. 11\,560--11\,569.

\bibitem{45}
J.~Cao, H.~Cholakkal, R.~M. Anwer, F.~S. Khan, Y.~Pang, and L.~Shao, ``D2det:
  Towards high quality object detection and instance segmentation,'' in
  \emph{2020 IEEE/CVF Conference on Computer Vision and Pattern Recognition
  (CVPR)}, 2020, pp. 11\,482--11\,491.

\bibitem{46}
N.~Carion, F.~Massa, G.~Synnaeve, N.~Usunier, A.~Kirillov, and S.~Zagoruyko,
  ``End-to-end object detection with transformers,'' in \emph{Computer Vision
  -- ECCV 2020}, A.~Vedaldi, H.~Bischof, T.~Brox, and J.-M. Frahm, Eds.\hskip
  1em plus 0.5em minus 0.4em\relax Cham: Springer International Publishing,
  2020, pp. 213--229.

\bibitem{47}
C.~Dong, C.~C. Loy, and X.~Tang, ``Accelerating the super-resolution
  convolutional neural network,'' in \emph{Computer Vision -- ECCV 2016},
  B.~Leibe, J.~Matas, N.~Sebe, and M.~Welling, Eds.\hskip 1em plus 0.5em minus
  0.4em\relax Cham: Springer International Publishing, 2016, pp. 391--407.

\bibitem{48}
C.~Dong, C.~C. Loy, K.~He, and X.~Tang, ``Image super-resolution using deep
  convolutional networks,'' \emph{IEEE Transactions on Pattern Analysis and
  Machine Intelligence}, vol.~38, no.~2, pp. 295--307, 2016.

\bibitem{50}
C.~Feng, Y.~Zhong, Y.~Gao, M.~R. Scott, and W.~Huang, ``Tood: Task-aligned
  one-stage object detection,'' in \emph{2021 IEEE/CVF International Conference
  on Computer Vision (ICCV)}, 2021, pp. 3490--3499.

\bibitem{23}
\BIBentryALTinterwordspacing
G.~Jocher, A.~Stoken, J.~Borovec, NanoCode012, ChristopherSTAN, L.~Changyu,
  Laughing, tkianai, A.~Hogan, lorenzomammana, yxNONG, AlexWang1900,
  L.~Diaconu, Marc, wanghaoyang0106, ml5ah, Doug, F.~Ingham, Frederik, Guilhen,
  Hatovix, J.~Poznanski, J.~Fang, L.~Yu, changyu98, M.~Wang, N.~Gupta,
  O.~Akhtar, PetrDvoracek, and P.~Rai, ``{ultralytics/yolov5: v3.1 - Bug Fixes
  and Performance Improvements},'' Oct. 2020. [Online]. Available:
  \url{https://doi.org/10.5281/zenodo.4154370}
\BIBentrySTDinterwordspacing

\bibitem{54}
C.~Zhang, Y.~Wu, M.~Guo, and X.~Deng, ``Training sample selection for
  space–time adaptive processing based on multi-frames,'' \emph{The Journal
  of Engineering}, vol. 2019, 10 2019.

\bibitem{55}
C.~H. Nguyen, T.~C. Nguyen, T.~N. Tang, and N.~L.~H. Phan, ``Improving object
  detection by label assignment distillation,'' in \emph{2022 IEEE/CVF Winter
  Conference on Applications of Computer Vision (WACV)}, 2022, pp. 1322--1331.

\end{thebibliography}
}

\end{document}